%% file: iclr2025_conference.tex
\documentclass{article} % For LaTeX2e
\usepackage{iclr2025_conference,times}

\input{math_commands.tex}

\usepackage{hyperref}
\usepackage{url}
\usepackage{amsmath}
\usepackage{amssymb}
\usepackage{mathtools}
\usepackage{amsthm}
\usepackage{fancybox}
\usepackage{CJKutf8}
\usepackage{multirow}
\usepackage{booktabs}

\usepackage{hyperref}
\usepackage{url}
\usepackage{enumerate}
\usepackage{algorithmic}
\usepackage{algorithm}
\usepackage{wrapfig}
\usepackage{float}

\newcommand{\mycomment}[1]{\textcolor{blue}{\textbf{#1}}} % 自定义注释格式

\title{Innovative Thinking, Infinite Humor: Humor Research of Large Language Models through Structured Thought Leaps}

\author{Han Wang\thanks{Equal Contribution. Work done during internship at Tencent QQ, as a part of QQ MLLM project} \ $^{1,2}$, Yilin Zhao$^{\textasteriskcentered 2}$, Dian Li\thanks{corresponding author} \  \thanks{Project leader of QQ MLLM project} \ $^2$, Xiaohan Wang$^{1,2}$, Gang Liu$^2$, Xuguang Lan$^{\dagger}$ \ $^1$, \\
	\textbf{Hui Wang}$^2$  \\
	Xi'an Jiaotong University$^1$; Tencent QQ$^2$ \\
	\texttt{reload7@stu.xjtu.edu.cn} ,\\ 
	 \texttt{\{yilinnzhao,goodli,shawnbywang,sinbadliu\}@tencent.com},\\
	  \texttt{xglan@mail.xjtu.edu.cn}, \texttt{joltwang@tencent.com}
}
\iclrfinalcopy

\begin{document}

\maketitle

\begin{abstract}
Humor is previously regarded as a gift exclusive to humans for the following reasons. Humor is a culturally nuanced aspect of human language, presenting challenges for its understanding and generation.
Humor generation necessitates a multi-hop reasoning process, with each hop founded on proper rationales. Although many studies, such as those related to GPT-o1, focus on logical reasoning with reflection and correction, they still fall short in humor generation. Due to the sparsity of the knowledge graph in creative thinking, it is arduous to achieve multi-hop reasoning.
Consequently, in this paper, we propose a more robust framework for addressing the humor reasoning task, named LoL. LoL aims to inject external information to mitigate the sparsity of the knowledge graph, thereby enabling multi-hop reasoning. In the first stage of LoL, we put forward an automatic instruction-evolution method to incorporate the deeper and broader thinking processes underlying humor.
Judgment-oriented instructions are devised to enhance the model's judgment capability, dynamically supplementing and updating the sparse knowledge graph. Subsequently, through reinforcement learning, the reasoning logic for each online-generated response is extracted using GPT-4o. In this process, external knowledge is re-introduced to aid the model in logical reasoning and the learning of human preferences.
Finally, experimental results indicate that the combination of these two processes can enhance both the model's judgment ability and its generative capacity.
These findings deepen our comprehension of the creative capabilities of large language models (LLMs) and offer approaches to boost LLMs' creative abilities for cross-domain innovative applications.
\end{abstract}

\section{Introduction}

\begin{figure*}[h]
	\begin{center}
		\includegraphics[width=0.8\textwidth]{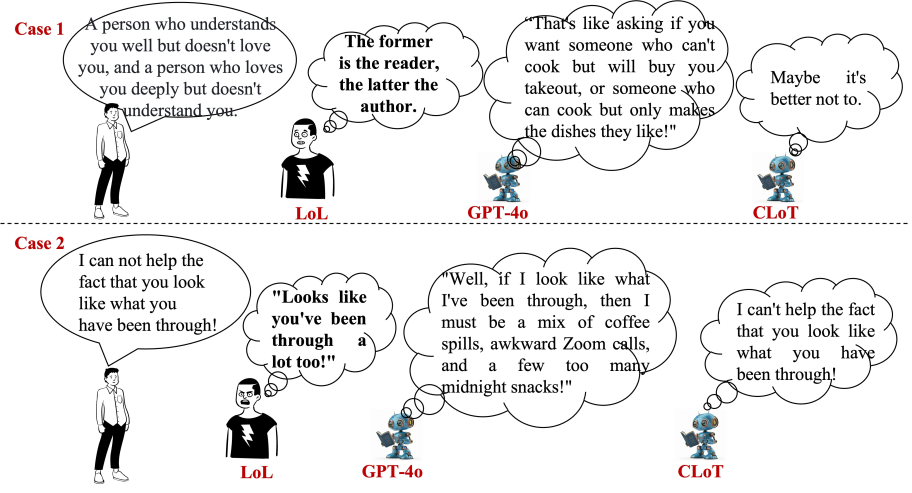}
	\end{center}
	\caption{English comparison showcase (more showcases are in Appendix \ref{showcase}). Compared to GPT-4o and CLoT, LoL provides shorter and more conversational answers to questions. For instance in Case 2, while LoL and CLoT may convey the same meaning, their different expressions produce different effects. Brief responses leave room for readers to ponder, enhancing interest and interactivity.}
	\label{exp11}
\end{figure*}

Currently, humor is attractive because, as shown in Figure \ref{exp11}, it requires a burst of inspiration, which is still difficult for humans.
The difficulty lies in the multi-hop reasoning process that fosters creativity. Each hop in this process is based on proper rationales. Without an understanding of these rationales, it is difficult for the model to grasp the internal humorous logic, making it prone to relying on pattern recognition.

% Creative thinking's difficulty lies in the sparsity of the knowledge graph behind it.  The sparsity is resulted from that entities in the  sparse knowledge graph (SKG) are far from each other and the relations in it is uncommon, which is further resulted from information missing.
So far, the Creative Leap of Thought (CLoT) paradigm \citep{xu2024good} has developed two basic abilities to facilitate humor generation: selection skill and ranking skill. With these two basic skills, CLoT introduces nouns as instruction back-translation for self-improvement. However, only question-answer pairs, i.e., the beginning and ending of the multi-hop reasoning path, are utilized to train the model. As mentioned in CLoT, this process only captures the inherent creative patterns within the data which impairs the generalization ability and fails to stimulate “thinking outside the box” for generating novel ideas.
Large language models (LLMs) such as GPT-4o or o1 \citep{lightman2023let} and QwQ \citep{qwq-32b-preview}, which show superior performance in almost all reasoning tasks, do not perform exceptionally well in humor generation, as shown in Figure \ref{exp11}. Moreover, humor-related works \citep{xu2024exploring, xu2024good} focus on a specific aspect of humor, like puns or proverbs, while humor also encompasses elements such as irony, limiting the range in real-world  applications.

For the multi-hop humor reasoning problem, understanding is fundamental for endowing LLMs with reasoning ability to avoid getting trapped in memorizing patterns. The introduction and augmentation of external knowledge help LLMs understand the underlying logic and rationale. Additionally, a reward model can optimize the behavior of large language models by providing feedback, enabling them to produce outputs that better meet expectations. Due to the subjectivity of humor, a unified score may contain significant noise. Thus, the judgment skill is essential for providing feedback to further enhance LLMs' reasoning ability. Finally, with these two basic skills, humor understanding and judgment, the ability to generate humor can be improved.

Therefore, we propose a more robust framework named LoL to address the humor reasoning task that current LLMs find challenging. LoL consists of two-stage training: the supervised Fine-Tuning (SFT) stage and the Direct Preference Optimization (DPO) stage. In the first stage, we develop human-designed judgment-related instructions and their derivatives to train the model's humor judgment capabilities. Additionally, we propose an automatic instruction expansion method for humorous conversations to inject and augment knowledge into the original training data, mimicking the human thinking process step by step. This will help LLMs deepen and broaden their understanding of humor content.
In the second stage, the reasoning rationale for each online-generated response is extracted using GPT-4o. In this process, external knowledge is introduced again to assist the model in logical reasoning and learning human preferences. The judgment capability from the first stage can also be useful for expanding the preference-pair dataset and further supplementing rationales.

We evaluated the humor judgment abilities of various large language models on both Chinese and English humor datasets. Experiments demonstrate that LoL outperforms other models on almost all test sets. Additional confirmatory experiments were conducted to show that LoL enhances the model's divergent thinking ability and effectiveness in humor generation. Our contributions are summarized as follows.
\begin{enumerate}
	\item[1.] We propose an automatic instruction-evolution system for conversation data. A three-agent system is introduced to inject and augment knowledge into the original training data. This will facilitate LLMs in deepening and broadening their understanding of the underlying logic and rationales.
	\item[2.] We propose a teacher-student prompt system to enhance the judgment ability of LLMs. Through the automatic construction of conversation data between the teacher and the student, LLMs learn the teacher's judgment of the student's thinking.
	\item[3.] Experimental results demonstrate that we can enhance both the model's judgment and generative capabilities and achieve state-of-the-art performance.
\end{enumerate}
%\vspace{-0.65cm}

\section{Method}
\label{method}

\subsection{Problem Formulation}
\label{sec-problem-formulation}

The multi-hop reasoning issue can be framed as a knowledge-graph (KG) problem. In this context, nodes (or concepts) along the multi-hop path are entities within the KG, and the rationales for enabling multi-hop reasoning are relations within the KG. By enriching the entities and relations in the KG, we can more easily explore the self-evolved path in multi-hop reasoning.

In general, the knowledge graph \( \mathcal{G} \) is defined as a set of triples \( \mathcal{G} = \{ (e, r, e') \mid e, e' \in \mathcal{E},\ r \in \mathcal{R} \} \), where \( \mathcal{E} \) is the set of entities and $\mathcal{R}$ is the set of relations. Each triple represents a relation $r$ from the head entity $e$ to the tail entity $e'$ \citep{sun2023think,yang2023knowledge}. In the specific application of humor generation, we consider a knowledge graph composed of question-related entities $\mathcal{E}_Q$ and answer-related entities $\mathcal{E}_A$. The intersection $\mathcal{E}_Z=\mathcal{E}_Q\cap\mathcal{E}_A$ is regarded as the set of correlation entities (we refer to them as correlation entities here).
Given the creative and unexpected nature of humor, as well as the existence of causal relationships between questions and answers, $\mathcal{E}_Z$ may consist of pseudo-correlation entities between $\mathcal{E}_Q$ and $\mathcal{E}_A$, such as those involved in puns (i.e., $\mathcal{E}_Z\rightarrow\mathcal{E}_Q$, $\mathcal{E}_Z\rightarrow\mathcal{E}_A$, and $\mathcal{E}_Q\rightarrow\mathcal{E}_A$). Additionally, it can also follow the pattern $\mathcal{E}_Q\rightarrow\mathcal{E}_Z\rightarrow\mathcal{E}_A$, as shown in Figure \ref{exp_back_front}. Clearly, it can be inferred that $\mathcal{E}_Z$ is pivotal for enabling the multi-hop in the humor reasoning.

Therefore, we formulate the causal relation $R_c$ into a verbal description as shown in Figure \ref{exm-rationale}, which contains the correlation entities $\mathcal{E}_Z$ either explicitly or implicitly. Finally, our objective is to expand the scopes of $\mathcal{E}_Q$ and $\mathcal{E}_A$, and utilize the causal relationship to structure the reasoning path. This is beneficial for mitigating the information insufficiency problem and further enabling humor reasoning.

The overall training framework is illustrated in Figure \ref{network}. In the first stage, supervised fine-tuning (SFT), we randomly initialize a LoRA model and train it using both single-turn and multi-turn question-answer format data. In the second stage, Direct Preference Learning (DPO), the model from the first stage serves as a reference model and is frozen to act as the judgment model when self-evolving. The tunable model is trained with preference question-answer data, which helps the model improve its logical reasoning and learn human preferences.

\subsection{Diverse Instruction Expansion and Tuning} 
\label{sec-diet}

A reward model can optimize the behavior of large language models by providing feedback, enabling them to generate outputs that better meet expectations. However, due to the subjectivity inherent in humor, a unified scoring system (i.e., a pointwise reward model) may be plagued by significant noise, such as topics, background and so on. Additionally, human-voting (i.e., Likes) data from the community is readily available. Transforming it into pairwise data is not only easier but also more in line with human intuition of selecting the more humorous one from two responses.
Consequently, we adopt a judgement model (i.e., pairwise reward model) to offer feedback, thereby further enhancing the reasoning capabilities of large language models (LLMs).

\begin{figure*}[th]
	\centering
	\includegraphics[width=0.95\textwidth]{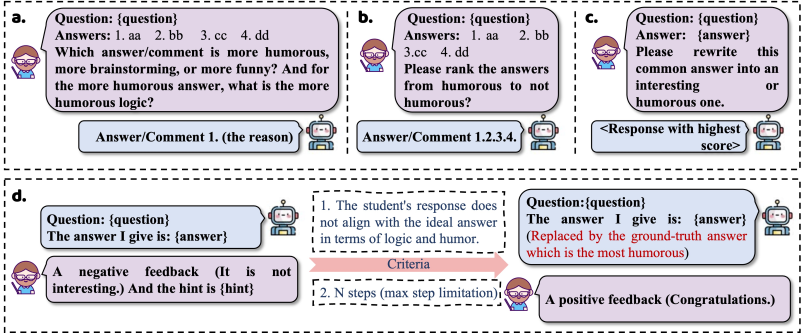}
	\caption{The details of judgement-oriented instructions templates.}
	\label{task_evol}
\end{figure*}

To stimulate the judgment capability of a model, we manually design a judgment-oriented template as shown in Figure \ref{task_evol}. Additionally, the understanding capability is enhanced to improve both the judgment and generation capabilities through automatic instruction evolution.

\textbf{Judgement Template Design.}
Question-answer data with human-voting annotations are utilized for judgment tasks. 

(1) \textit{Selection and Ranking Skill.} Two basic templates are designed, as shown in Figure \ref{task_evol} (a) and (b). Single-choice and ranking questions are basic tasks for developing judgment capabilities.
Furthermore, to deepen the model's understanding of the contrast between "humorous" and "non-humorous", two additional tasks are proposed as shown in Figure \ref{task_evol} (c) and (d).

(2) \textit{Answer Rewriting} (template in Figure \ref{task_evol}(c)). We identify the answer receiving the most human votes as the most humorous one. Subsequently, GPT-4o is employed to rewrite this most humorous answer into a non-humorous version. Finally, the model is trained to transform a non-humorous statement into a humorous expression using single-round session data format.

(3) \textit{Teacher-Student Prompt Loop} (template in Figure \ref{task_evol}(d)).
Similar to the human thought process, the complete exploration process encompasses trial-and-error, reflection, and backtracking. Within this process, the ability to make sound judgments in a long chain of thought is crucial. Therefore, a guided conversation process between two agents is proposed to enhance the reflection, and backtracking.
We utilize two GPT-4o models and designate one as the teacher and the other as the student. The teacher provides a judgment conclusion and provides prompts based on the given question and the ideal answer. The student generates an answer using the teacher's prompts and the original question. Two criteria as shown in Figure \ref{task_evol}(d) are applied to stop this multi-turn conversation.

\begin{figure*}[ht]
	\begin{center}
		%\framebox[4.0in]{$\;$}
		\includegraphics[width=0.95\textwidth]{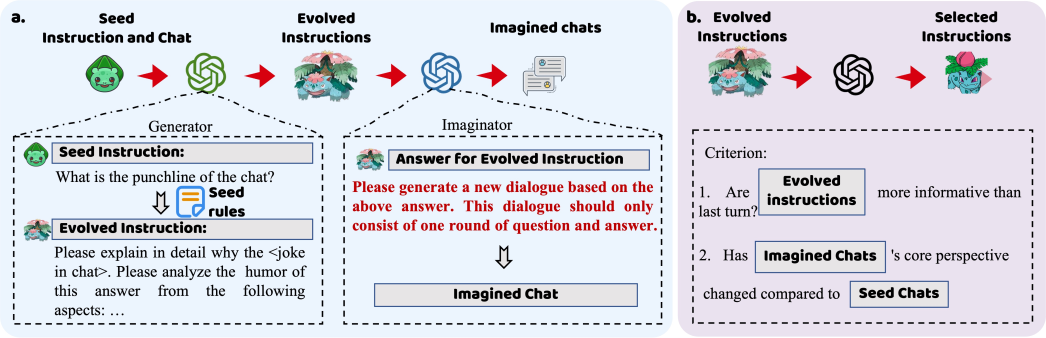}
	\end{center}
	\caption{The detail of AIE. And the detailed process is shown in Algorithm \ref{alg-aaie}}
	\label{img-aaie}
\end{figure*}

\textbf{Automatic Instruction Expansion.}
In the human cognitive process, mechanisms such as trial and error, reflection, and backtracking are inherently supported by a robust understanding of concepts. To emulate the deep and extensive understanding characteristic of human cognition, we propose a three-agent system inspired by automatic instruction evolution, which leverages the extensive world knowledge encapsulated in large language models (LLMs) to autonomously enrich prompts within the realm of humor comprehension. This methodology not only facilitates the exploration of unconventional associations among disparate concepts but also enhances the model's ability to establish and strengthen interconnections between diverse ideas. The process is shown in the Figure \ref{img-aaie}.

% Given a seed conversation dataset $D = \{ (q_k^0, a_k^0) \}_{k = 0}^{N}$, where $q_k^0$ is a question, $a_k^0$ is a funny answer, and $N$ is the number of samples. Let $I=\{I_k | I_k=(q_k^0,a_k^0, i_k^0)\}_{k = 0}^N$ be the prompts to be expanded, where $i_k^0$ is an instruction for understanding the conversation $(q_k^0,a_k^0)$. Let $y_k^0 \in Y_k$ be the reply to $i_k^0$.
% The framework in Figure \ref{img-aaie} depicts the process of instruction expansion for knowledge augmentation and involves three agents, each assuming the roles of \textit{generator}, \textit{selector}. Here's how the process unfolds:

Given a seed conversation dataset $D = \{ (q_k^0, a_k^0) \}_{k=0}^{N}$ where $q_k^0$ denotes questions and $a_k^0$ represents humorous responses. Let $I_0 = \{I_k | I_k=(q_k^0,a_k^0, i_k^0)\}_{k=0}^N$ be the initial instruction set, where each instruction $i_k^0$ contextualizes the dialogue pair $(q_k^0,a_k^0)$. The framework implements a 3-round evolutionary process through three cooperative agents.
The iterative process terminates when either Criterion-2 is satisfied or maximum evolution rounds ($t \geq 3$ in our setting)  are reached. Each iteration produces enhanced instruction-output pairs $(I_{new}, y_{new})$ where $y_{new} \in Y$ denotes optimized responses under the augmented instruction space.

\textit{Generation}:
The Generators module streamlines the process of instruction generation, enabling the automated generation of instruction data based on a seed conversation. Unlike evol-instruct methods in math and other fields, we design an "imaginator" in the generation process for humor reasoning. It tries to guess the seed conversation based on the answer from evolutionary instruction. This can cause a shift in the storyline within the conversation topic. In other words, it tries to explore the boundaries of the knowledge graph under the context of the seed conversation and enhance creativity.

\textit{Selection}:
The Selectors module is crafted to streamline the instruction-filtering process, enabling the curation of instruction datasets from evolved instruction data. In contrast to evol-instruct techniques in mathematics and other domains, this approach serves not merely as a means to unleash the capabilities of LLMs, but also to boost the capacity for thinking outside the box, thereby further actualizing creativity. Therefore, we design a critic related to the conversation topic, which limits the model's thinking from being overly divergent.

Finally, the system outputs multi-turn question-answer data which format is $\{(q_k^0, a_k^0), (I_k^0, y_k^0), \dots, (q_k^{m_k}, a_k^{m_k}), (I_k^{m_k}, y_k^{m_k})\}_{k=0}^N$, where $m_k$ is the maximum number of communication rounds between the three agents.
Finally, all above human-designed and automatic expanding data are involved in training.

\subsection{Guide Exploration and Self-Improvement Tuning}
\label{sec-gesit}

\begin{figure*}[th]
	\begin{center}
		%\framebox[4.0in]{$\;$}
		\includegraphics[width=0.95\textwidth]{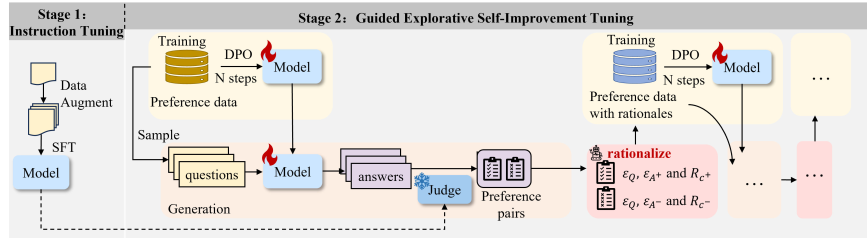}
	\end{center}
	\caption{Pipeline of training process.}
	\label{network}
\end{figure*}

In the previous section, we developed a model with judgment and understanding capabilities. Through multiple rounds of dialogue, we strengthened the relationships among diversified entities $\mathcal{E}_Q$ and $\mathcal{E}_A$.
At this stage, the policy model enhances its performance through online self-learning, using a dataset of (question, humorous answer, non-humorous answer) tuples: $(q, a^+, a^-)$. Specifically, as shown in Figure \ref{network}, it learns by repeatedly sampling responses, evaluating answer correctness with the capability from Section \ref{sec-diet}, and updating its parameters online using Direct Preference Optimization \citep{rafailov2024direct}.

As detailed in  Algorithm \ref{gest}, we start with an initial offline preference dataset without rationales, \(D_0 = \{(q_i, a_i^+, a_i^-)\}_{i = 0}^M\), where \(M\) is the number of training data. The model trained in Section \ref{sec-diet} is copied into two. One, denoted as \(\pi^*\), is frozen to judge whether sampled answers are humorous for online data augmentation. The other, \(\pi\), is trained to have a more robust generative ability.

During training, the policy model \(\pi\) trains on \(D_0\) for a few steps. Every \( T \) steps, \(\pi\) selects \(l\) questions from \(D_0\) to sample two answers or each question. The judge model \(\pi^*\) then determines the more humorous one, constructing new pairwise data incorporated into the training dataset. Additionally, a causal inference expert (GPT-4o) provides rationale for each answer relative to the question (see Appendix \ref{apd-GESIT} for details). This process expands the preference dataset to \(D = D_0 \cup \widetilde{D}\), where \(\widetilde{D} = \{ (q, \widetilde{a}^+, \widetilde{a}^-, \widetilde{r}^+, \widetilde{r}^-) \}\).

\section{Experiments}
\label{Experiments}

We collect humor-related data with human-voting from humor games and communities, and then organize it into the format mentioned in Section \ref{sec-diet}.
Since we enhance the generation ability based on the judgment capability, it is crucial to verify the performance in humor-judgment tasks. Thus, we construct single-choice questions at different difficulty levels and carry out the consensus generation task validation.

\subsection{Datasets source}
\label{data}

(1) Oogiri-GO \citep{zhong2024let}. 
In the "Oogiri game", participants are required to give unexpected and humorous responses to given images, text, or both. This game demands a sudden burst of insight and strong associative thinking within the given context. Similar to the processing method of CLoT \citep{zhong2024let}, we randomly select 95\% of the samples to construct the training dataset, and the remaining 5\% is used to form the test dataset for validation and analysis.

(2) SemEval 2021 Task 7 \cite{meaney2021semeval} contains binary labels and ratings collected from a balanced age group ranging from 18 to 70 years old. Data with binary labels and ratings are utilized to construct the easy-case task and the hard-case task mentioned in Section \ref{se-tasks} respectively.

(3) SemEval 2020 Task 7 \citep{hossain-etal-2020-semeval} is a game to change a word in a headline to make it funnier. It also contains binary labels and ratings which are utilized to construct the easy-case task and the hard-case task.

(4) Chinese Community Data. We collect data from various Chinese communities, including Ruozhiba and others. Human-voting (i.e., Likes) labels are readily available, and we use them to construct both easy-case and hard-case tasks.

\subsection{Tasks construction} 
\label{se-tasks}

(1) Inspired by the task design in CLoT\citep{zhong2024let}, we develop judgement-related tasks as follows.

\textbf{Easy-case Task (i.e. $2T1$).} Single-choice-from-two-options questions are constructed from the binary-label data mentioned above and the data with the largest gap in human-voting counts. The construction adheres to the template depicted in Figure \ref{task_evol}(a).

\textbf{Hard-case Task (i.e. $2T1(Hard), 3T1, 4T1$).} Single-choice-from-two(or three or four)-options questions are constructed from the same source with a closer gap in human-voting counts. The construction adheres to the template depicted in Figure \ref{task_evol}(a).

\begin{wraptable}{r}{6cm}
	\centering
	\caption{Humor judgement validation on ruozhiba dataset}
	\label{}
    \renewcommand\arraystretch{1.2}
	\begin{tabular}{c|c}  
		\toprule[1pt]
            Model & 2T1 \\
            \hline
		GPT-4o & 76.40  \\ 
		\hline
            QWEN1.5-32B &68.85 \\
            \hline
            CLoT &50.20 \\
            \hline
            QwQ &90.80 \\
            \hline
            LoL & 95.35  \\ 
            \bottomrule[1pt]
    \end{tabular}
\end{wraptable}

(2) Inspired by the generation tasks designed in CLoT\citep{zhong2024let}, the Divergent Associate Thinking (DAT) task \citep{olson2021naming} to validate the capability of creativity is carried out. Additionally, human evaluation will also be validated.

\subsection{Results Analysis}

\textbf{Evaluation on Judgement tasks in English.} 
We validate the top-1 accuracy of completing each judgement task and show the performance of several models in Table \ref{en-bench}. Overall, compared with open-source language models such as LLaMA3, QWEN and so on, state-of-the-art closed-source LLMs exhibit impressive zero-shot performance in humor judgement tasks. Model trained by LoL has significantly improved compared to other models (such as LLAMA3-70B and GPT-4o). Specifically, the average accuracy in diverse English benchmarks has increased by 4.55\% and 5.91\% respectively.

\textbf{Evaluation on Judgement tasks in Chinese.} 
We also evaluate the accuracy rate (acc\%) of completing each selection task in Chinese and show the performance of several models in Table \ref{ch-bench}. Overall, compared with open-source language models including LLaMA3 and QWEN, the state-of-the-art closed-source large language models show impressive zero-shot performance on humor judgement tasks in Chinese. The model trained by LoL also shows a significant improvement compared to other models (such as GPT-4o) (with an average accuracy in diverse Chinese benchmark increase of 16.22\%).

We also conduct an additional experiment on the Ruozhiba dataset\footnote{https://github.com/Leymore/ruozhiba/tree/main?tab=readme-ov-file}, which most well-known large language models (LLMs) have been trained on. We asked GPT-4o to rewrite the Ruozhiba queries into question-answer pairs, placing the punchline in the answer section. Then, we asked GPT-4o again to rewrite the ground-truth answers into non - humorous versions. Based on the positive-negative pair data, the LLMs were tested, and the results are shown in Table 9. The results indicate that CLoST also achieves state-of-the-art performance on the Ruozhiba dataset.

\begin{wraptable}{l}{7cm}
	\caption{The accuracy (\%) of choice questions on various Algorithms in Chinese benchmarks.}
	\label{ch-bench}
	\renewcommand\arraystretch{1.2}
	\begin{tabular}{c c|c c}
		\toprule[1pt]
		\multicolumn{2}{c|}{\multirow{2}*{Model}} & \multicolumn{2}{c}{Chinese Benchmark}\\
		\cline{3-4}
		& &2T1&2T1(hard)\\
		\cline{1-4}
		\multirow{1}*{GPT}&4o&64.98 & 63.49  \\
		\cline{1-4}
		\multirow{2}*{LLAMA3}&8B& 50.72 & 57.44   \\
		%			\cline{3-9}
		\multicolumn{1}{c}{}&70B& 59.48 & 61.22\\
		\cline{1-4}
		\multirow{3}*{QWEN1.5}&7B&54.82 & 51.71 \\
		%			\cline{3-9}
		\multicolumn{1}{c}{}&14B&53.45 & 57.41\\
		\multicolumn{1}{c}{}&32B&52.71 & 56.27 \\
		\cline{1-4}
		\multirow{2}*{QWEN2}&7B& 51.99 &  58.17 \\
		%			\cline{3-9}
		\multicolumn{1}{c}{}&57B& 65.91 & 57.03 \\
		\cline{1-4}
		\multirow{1}*{QWEN2.5}{}&32B& 61.53 & 60.46 \\ 
		\cline{1-4}
		\multirow{1}*{Baichuan2}{}&13B& 50.56 & 53.61 \\ 
		\cline{1-4}
		\multirow{1}*{CLoT}{}&7B &52.12 & 34.46\\ 
		%			\hline
		\cline{1-4}
		\multirow{1}*{QwQ}{}& 32B&59.75 & 57.04\\ 
		%			\hline
		\cline{1-4}
		\multirow{1}*{QwQ}{+LoL}& 32B&88.91 & 63.12\\ 
		\cline{1-4}
		\multirow{1}*{OURS}&32B&\textbf{90.95} & \textbf{69.97}\\
		\bottomrule[1pt]
	\end{tabular}
\end{wraptable}

\begin{figure}[th]
	\begin{minipage}{0.23\linewidth}
		%		\vspace{1pt}
		\centerline{\includegraphics[width=\textwidth]{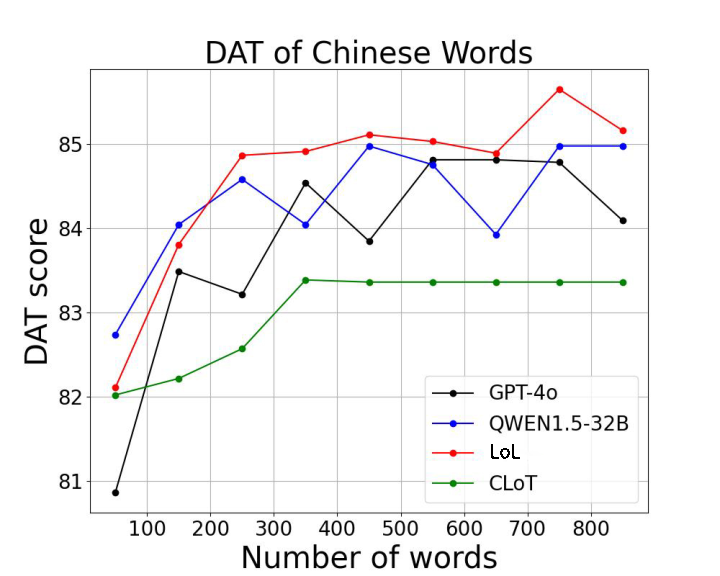}}
		\centerline{(a)}
	\end{minipage}
	\begin{minipage}{0.23\linewidth}
		%		\vspace{1pt}
		\centerline{\includegraphics[width=\textwidth]{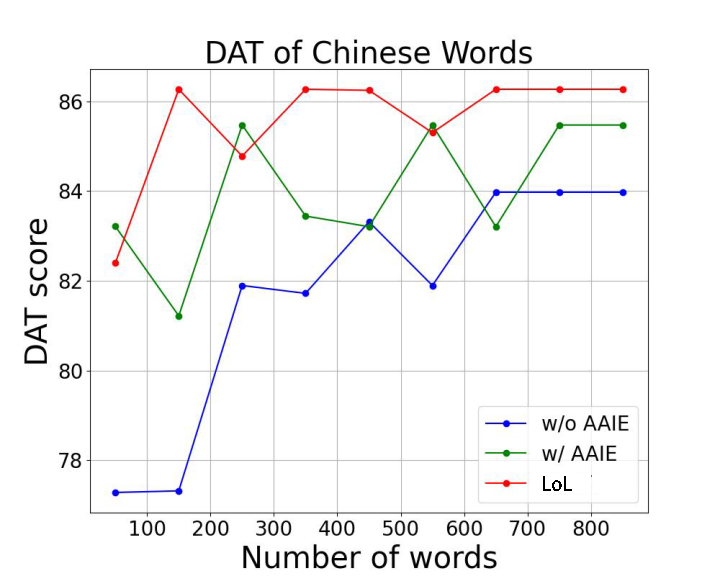}}
		\centerline{(b)}
	\end{minipage}
	\begin{minipage}{0.23\linewidth}
		%		\vspace{1pt}
		\centerline{\includegraphics[width=\textwidth]{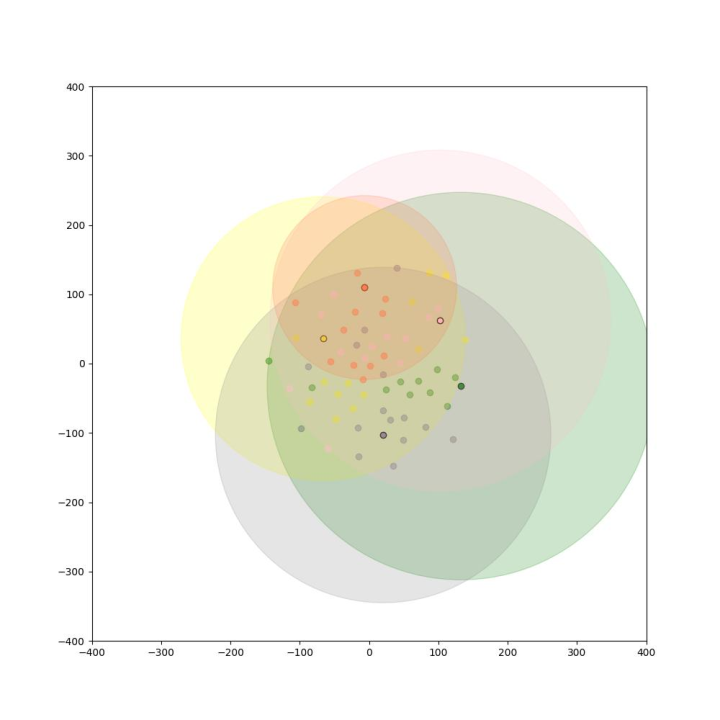}}
		\centerline{(c)}
	\end{minipage}
	\begin{minipage}{0.23\linewidth}
		%		\vspace{1pt}
		\centerline{\includegraphics[width=\textwidth]{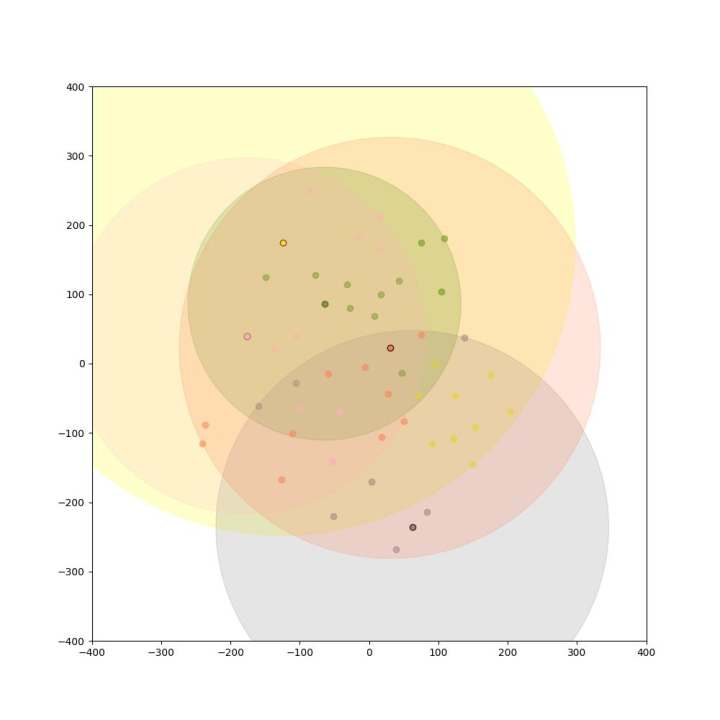}}
		\centerline{(d)}
	\end{minipage}
	\caption{Divergent associate thinking (DAT) validate. (a). DAT score compared among LoL and three baselines (b). DAT score compared among different component of LoL  (c). TSNE Results of Word Vectors Obtained by QWEN1.5-32B Associating Five Target Words. (d). TSNE Results of Word Vectors Obtained by LoL Associating Five Target Words.}
	\label{fig4}
\end{figure}

\textbf{Evaluation on Creative-thinking validation in Generation Tasks.} 
To evaluate the associative generalization capability of LoL, we test it on a creative task known as the Divergent Association Task (DAT). The DAT is a classic creativity test in which testee write 10 unrelated words, and words with greater “distances” in their context embeddings indicate more divergent thinking. In the Chinese creativity test, we utilize Chinese Word Vectors \citep{li2018analogical} to calculate the DAT score. First, we provide specific words and ask the model to generate associations and imaginations, obtaining 10 associated words. Then, we use these 11 words to calculate the DAT score. Additionally, we test how the model's DAT score varies as the number of test words increases. It can be observed that as the number of words increases, the DAT score also tends to stabilize. From the average of each domain, it can be seen that LoL has the highest score after stabilization. As shown in Figure \ref{fig4}, LoL has a slight performance improvement in the mean value of DAT compared to Qwen1.5-32B, GPT-4o and CLoT.

In addition, we employ T-SNE to project the embedding vectors of these words into a two-dimensional space. specifically, the target word is positioned at the center, and a circle is drawn with a radius equal to the Euclidean distance between the target word's embedding and that of the farthest associated word. In Figure \ref{fig4}(c) and (d), the embedding vectors of five target words and their associated words are illustrated, with different colors representing different target vectors. A larger circle indicates a broader semantic range for the target word, implying a greater number of associated words. In both tests, LoL outperforms previous works including SOTA models like GPT-4o.

\begin{table*}[t]
	\begin{small}
		\begin{sc}
	\caption{The accuracy (\%) of choice questions on various Algorithms in English benchmarks.}
	\label{en-bench}
	\begin{center}
		\renewcommand\arraystretch{1.2}
		\begin{tabular}{c c|c c c c|c|c}
			\toprule[1pt]\multicolumn{2}{c|}{\multirow{2}*{Model}} & \multicolumn{4}{c|}{SemEval 2021}& \multicolumn{1}{c|}{SemEval 2020}& \multicolumn{1}{c}{Oogiri-GO}\\
			\cline{3-8}
			& &2T1&2T1(hard)&3T1&4T1&2T1&2T1\\
			\cline{1-8}
			\multirow{1}*{GPT}&4o&85.09 & \textbf{60.77} & 43.71 & 34.63& 55.08 & 85.09 \\
			%			\cline{3-9}
			%			&\multicolumn{1}{c}{}&CLoT&52.49 & 51.74 & 34.46 & 23.59&0&0\\
			\cline{1-8}
			%			&\multirow{2}*{LLAMA2}&7B&43.50 & 35.33 & 0 & 0&0& \\
			%%			\cline{3-9}
			%			&\multicolumn{1}{c}{}&14B&0&0&0&0&0&0\\
			%			\cline{1-8}
			\multirow{2}*{LLAMA3}&8B& 43.85 & 54.23 & 39.81 & 29.57 & 59.93 & 72.05\\
			%			\cline{3-9}
			\multicolumn{1}{c}{}&70B& 93.60 &58.08&39.81& 31.82 & 60.73 & 88.51\\
			\cline{1-8}
			\multirow{3}*{QWEN1.5}&7B& 62.04 & 52.02 & 31.54 & 24.89 & 50.46 & 36.65 \\
			%			\cline{3-9}
			\multicolumn{1}{c}{}&14B& 82.05 & 51.04 & 30.92 & 24.24 & 50.38 & 53.73\\
			\multicolumn{1}{c}{}&32B& 68.01 & 52.57 & 35.38 & 28.79 & 56.39 & 68.01\\
			\cline{1-8}
			\multirow{2}*{QWEN2}&7B& 56.55 & 50.63 & 32.31 & 23.38 & 50.08 & 62.11  \\
			%			\cline{3-9}
			\multicolumn{1}{c}{}&57B& 83.30 & 52.02 & 37.08 & 28.79 & 48.29 & 48.14 \\
			\cline{1-8}
			\multirow{1}*{QWEN2.5}{}&32B& 94.00 & 55.22 & 34.77 & 27.92 & 58.71 & 81.68 \\ 
			\cline{1-8}
			\multirow{1}*{Baichuan2}{}&13B& 51.70 & 52.71 & 35.69 & 24.24 & 51.45 & 50.00 \\ 
			\cline{1-8}
			\multirow{1}*{CLoT}{}& 7B &52.49 & 51.74 & 34.46 & 23.59& 53.50 & 52.49\\ 
			%			\hline
			\cline{1-8}
            \multirow{1}*{QwQ}{}& 32B &80.05 & 53.06 & 33.49 & 24.66 & 56.58 & 59.63\\ 
            \cline{1-8}
			\multirow{1}*{OURS}&32B&\textbf{96.58} & 57.45 & \textbf{48.06} & \textbf{35.90} & \textbf{64.57} & \textbf{97.20}\\
			%			\cline{2-9}
			\bottomrule[1pt]
		\end{tabular}
	\end{center}
\end{sc}
\end{small}
\end{table*}

\begin{table*}[t]
		\begin{small}
		\begin{sc}
	\caption{Ablation on English benchmarks.}
	\label{en-bench-abla}
	\begin{center}
		\renewcommand\arraystretch{1.2}
		\begin{tabular}{p{3cm}| c c c c|c}
			\toprule[1pt]
			\multicolumn{1}{c|}{\multirow{2}*{Model}} & \multicolumn{4}{c|}{SemEval 2021} & Oogiri-GO-en  \\
			\cline{2-6}
			&2T1&2T1(hard)&3T1&4T1&2T1 \\
			\cline{1-6}
			\multirow{1}*{QWEN1.5}{-32B}&68.01 & 52.57 & 35.38 & 28.79 & 68.01    \\ 
			\cline{1-6}
			\multirow{1}*{Oogiri-GO}{+Figure \ref{task_evol}(c)}& 55.12 & 53.27  & 33.54 & 27.49 & 81.68   \\ 
			\cline{1-6}
			\multirow{1}*{Oogiri-GO}{+Figure \ref{task_evol}(c,d)}& 89.70 & 50.28 & 30.46 & 17.32 & 95.96   \\ 
			\cline{1-6}
			\multirow{1}*{Oogiri-GO}{+AIE} & 88.40 & 50.07 & 30.92 & 19.48 & 96.58 \\ 
			\cline{1-6}
			\multirow{1}*{ALL}{+Figure \ref{task_evol}(c,d)} & 92.25 & 52.81 & 37.46 & 30.26 & 96.27  \\ 
			\cline{1-6}
			\multirow{1}*{ALL}{+AIE(DIET)}& \textbf{94.25} & \textbf{53.10} & \textbf{45.46} & \textbf{32.26} & \textbf{97.20}  \\ 
			\bottomrule[1pt]
		\end{tabular}
	\end{center}
\end{sc}
\end{small}
\end{table*}

\begin{figure}[ht]
		\begin{minipage}{0.32\linewidth}
				%		\vspace{1pt}
				%这个图片路径替换成你的图片路径即可使用
				\centerline{\includegraphics[width=\textwidth]{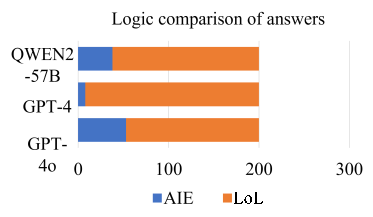}}
				% 加入对这列的图片说明
				\centerline{(a)}
			\end{minipage}
	\begin{minipage}{0.32\linewidth}
		%		\vspace{1pt}
		\centerline{\includegraphics[width=\textwidth]{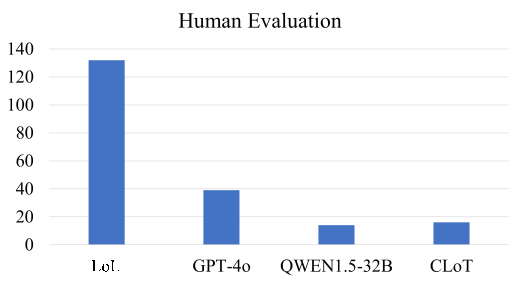}}
		\centerline{(b)}
	\end{minipage}
	\begin{minipage}{0.32\linewidth}
		%		\vspace{1pt}
		\centerline{\includegraphics[width=\textwidth]{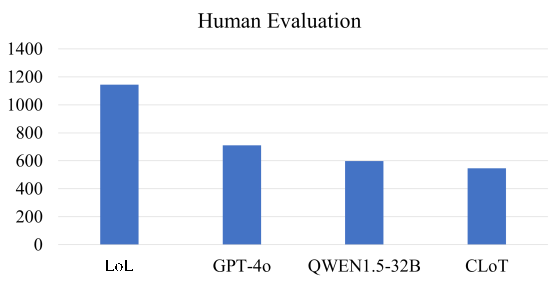}}
		\centerline{(c)}
	\end{minipage}
	\caption{(a) GPT-4  and GPT-4o logically evaluates the output of the model after DIET and LoL respectively. (b). Human evaluation about the win rate statistics based on the problem dimension. (c). Human evaluation about the win rate based on the total number of votes received by the four LLMs.} 
	\label{fig5}
\end{figure}

\textbf{Human Evaluation in Generation Tasks.} 
We conduct a user-preference study to test the creativity and humor of LLMs. Here, we select four LLMs (LoL, GPT-4o, QWEN1.5-32B, CLoT) to generate responses for a total of 200 text-based questions. We present a question and several corresponding replies from the four LLMs, and ask users to choose the most creative and humorous response. Figure \ref{fig5}(c) summarizes the statistical analysis of 3000 valid surveys. Figure \ref{fig5}(b) shows the win-rate calculated based on the problem dimension. The results indicate that users have a strong inclination to select the results of LoL, highlighting the high - quality creative content generated by CLoT. For more details of the user study, refer to the Appendix \ref{}.

\textbf{Ablation Study}

\begin{wraptable}{r}{6.6cm}
% \begin{table}
	\vskip 0.15in
	\centering
	\caption{Ablation on Chinese benchmarks.}
	\label{ch-bench-abl}
	\renewcommand\arraystretch{1.2}
	\begin{small}
	\begin{sc}
	\begin{tabular}{c|c c}
		\toprule[1pt]
		\multicolumn{1}{c|}{\multirow{2}*{Model}} & \multicolumn{2}{c}{Chinese Benchmark}\\
		\cline{2-3}
		&2T1&2T1(hard)\\
		\cline{1-3}
		\multirow{1}*{QWEN1.5}{-32B}& 52.71 & 56.27  \\
		\cline{1-3}
		\multirow{1}*{+Figure \ref{task_evol}(c)}& 63.56 & 63.88   \\
		\cline{1-3}
		\multirow{1}*{+Figure \ref{task_evol}(c)(d)}& 83.62 & \textbf{64.64}\\
		\cline{1-3}
		\multirow{1}*{+AIE (DIET)}& \textbf{90.34} & \textbf{64.64} \\
		\bottomrule[1pt]
	\end{tabular}
\end{sc}
\end{small}
% \vskip-0.1in
% \end{table} 
\end{wraptable}
We examine the ablation effects of different components in LoL and report the performance results in Table \ref{en-bench-abla} and Table \ref{ch-bench-abl}. In Table \ref{en-bench-abla}, line 1 presents the performance of QWEN-1.5-32B. Lines 2-4 show the performance of gradually adding methods on the Oogiri-GO-en dataset. The results indicate that with the increase in the number of tasks, especially in the teacher-student system, the judgment performance improves a lot. Employing AIE on the Oogiri-GO-en dataset only causes a slight decrease in performance. This might be caused by overfitting to the divergence of thought in the Oogiri-GO-en dataset. Lines 5-6 show that when all datasets with the introductory-part method are used to train the model, a further increase in performance is realized.

To evaluate the performance of GESIT, which primarily enhances the model’s causal and logical reasoning capabilities (i.e., the ability to relate to internal contexts), we enlisted three experts to assess the logical coherence of responses on 200 examples from GPT-4o, GPT-4 and QWEN2-57B, respectively. The experimental results shown in Figure \ref{fig5}(a) demonstrate that incorporating GESIT significantly strengthens the logical reasoning in the model’s answers.
In addition, the DAT test is conducted on ablation study in Figure 5(b), which shows that AIE enhance the divergent associate thinking ability. 

\section{Related Works}
% humor, reasoning and self evolve
\textbf{Large Language Models (LLMs) and Their Creativity.} Recently, language models \citep{bai2023qwen, wang2023cogvlm, liu2024improved, chen2023minigpt} have garnered widespread attention due to their impressive reasoning capabilities \citep{wang2023cogvlm,saparov2022language,zeng2022socratic,driess2023palm,dong2023dreamllm,ye2023mplug,liang2024toa}. Additionally, an increasing number of studies are focusing on exploring the creativity of LLMs \citep{ling2023unleashing,summers2023brainstorm,sun2023inspire,bhavya2023cam}, with applications in fields such as scientific discovery \citep{park2023papers,kang2022augmenting,hope2022scaling,liang2022stiffness,huang2023fast} and creative writing \citep{swanson2021story,chakrabarty2022help,wu2022promptchainer,mirowski2023co,dang2023choice}.

\textbf{Computational humor} is a branch of computational linguistics and artificial intelligence that utilizes computers to study humor \citep{binsted2006computational}. It encompasses various tasks, including humor discrimination \citep{shahaf2015inside,tanaka2022learning,xu2022hybrid,chen2022integrating,kumar2022deephumor,wu2021mumor,ofer2022cards,xie2023funqa,meaney2021semeval,hossain-etal-2020-semeval}, humor interpretation \citep{hwang2023memecap,evans2019gender,vasquez2021cats}, and humor generation \citep{amin2020survey,zhang2020let,hossain2020stimulating,valitutti2013let,chaudhary2021towards}. With the advancements in generative LLMs, humor generation has become a focal point due to its demand for creative thinking. \citep{zhong2024let} extends the chain-of-thought paradigm into humor generation. However, only question-answer pairs and nouns as bask-translation, i.e., the beginning and ending of the multi-hop reasoning path, are utilized to train the model, and this process only captures the inherent creative patterns within the data which impairs the generalization ability and fails to inspire “thinking outside the box” for generating novel ideas. Therefore, we develop a reasoning process featuring the processes of thinking divergence and reflection for humor generation.

\textbf{Instruction evolutionary.} Recent attention has focused on the complex instruction-following capabilities of Large Language Models (LLMs), leading to the development of new evaluation benchmarks \citep{zhou2023instruction, jiang2023followbench, qin2024infobench}. These studies consistently reveal that open-source LLMs lag behind proprietary models in their ability to follow complex instructions. However, there has been limited research on techniques to enhance this capability. Notable exceptions include Evol-Instruct \citep{xu2023wizardlm, zeng2024automatic} and Conifer \citep{sun2024conifer}, which encourage LLMs to evolve instruction complexity and apply supervised fine-tuning (SFT) on the data generated from this process. Automated generation is a widely-used method for Evol-Instruct, minimizing the reliance on extensive human annotation or manual data collection. This approach utilizes LLMs to create large volumes of instructional data sourced from chat data \citep{chiang2023vicuna} or by expanding a small set of seed instructions \citep{wang2022self, xu2023wizardlm, li2023self}. The generated instructions are then used to derive corresponding inputs and outputs. In this paper, LoL develops an automatic instruction evolution method with three agents to strengthen the ability of thinking divergence and humor understanding.

\textbf{Large language models reasoning.} Human cognition involves two distinct modes of processing: one that is fast and intuitive, and the other that is deliberate and analytical. Currently, LLMs can not only generate rapid responses using learned patterns, but more significantly, simulate complex reasoning processes through mechanisms like chain-of-thought \citep{wu2022promptchainer, wei2022chain,zhang2022automatic, yao2024tree, long2023large} or other forms of search, similar to how humans engage in deeper, step-by-step thinking \citep{fu2022complexity,lightman2023let, lai2024step}. However, these approaches build on existing LLMs without truly embedding the chain-of-thought ability within the models themselves. As a result, LLMs cannot inherently learn this reasoning capability, leading to active research on how to integrate it directly into model training. Most of these efforts\citep{qin2024o1, luong2024reft, qwq-32b-preview, qwen2, qvq-72b-preview, Qwen2VL, skyworkopeno12024} focus on improving LLM reasoning by integrating process supervision, reinforcement learning (RL), and inference-time computation strategies such as guided search. 
By doing so, it shifts the focus from merely scaling model parameters during pre-training to leveraging smarter inference strategies at test time. These techniques help the model refine its reasoning step by step, allowing it to pause, evaluate intermediate reasoning, and select better solution pathways during test-time computation. All of these are adept at factual reasoning like math or coding reasoning, while humor reasoning is mostly non-factual reasoning requiring creative thinking.

\section{Conclusion}

In this paper, we introduce the LoL method aimed at enhancing the generation capabilities of large language models (LLMs). LoL commences by transforming humor datasets into instruction-tuning data to train LLMs, thus improving their logical thinking (LoT) and judgment abilities. Subsequently, LoL utilizes Guided Explorative Self-Improvement Tuning, enabling LLMs to generate more creative structured thought data by understanding rationales and to select high-quality data for self-refinement training. Experimental results illustrate the effectiveness and generalization ability of LoL across diverse tasks, such as witty response generation and humor discrimination.


\subsubsection*{Acknowledgments}
This work was done during internship at Tencent QQ, as a part of QQ MLLM project and supported in part by NSFC under grant No.62125305, No.62088102, No. U23A20339, No. 62203348.

\bibliography{iclr2025_conference}
\bibliographystyle{iclr2025_conference}

\appendix
% \section{Appendix}
\section{Appendix}

\subsection{Training and Experiment Details}
\textbf{Training pipeline:}\\
LoL takes a two-stage training strategy. In the first process (supervised fine-tuning (SFT)), we randomly initialize a LoRA model. And we train the model with single-turn question-answer format data (data from Figure 2(a)(b)(c)) and muti-turn question-answer format data (data from Figure 2(d) and AIE). In the second process (Direct Preference Learning (DPO)), the first stage model serves as ref model, and it is frozen as judgement model. The tunable model is trained to improve the reasoning generation capability. At the beginning of stage 2, only preference question-answer data without rationale is fed into the tunable model for training.  After several steps, the rationale for each online generated response is extracted using GPT-4o and the preference question-answer data with rationles are mixed into the original dataset. And in each batch, the ratio of 'w' and 'w/o' rationale is $1 : 1$. 

\textbf{Implementation Details}
We validate the validity of LoL mainly based on QWEN1.5-32B-Chat \cite{bai2023qwen} using the LoRA \cite{hu2021lora} on 8 A100 GPUs.
For the first stage, we train the model on 95\% of the dataset mentioned above for 6 epochs using the AdamW optimizer with a learning rate of $3e-4$. In the second stage, 5\% of the dataset is used to train GESIT for 3 epochs using the AdamW optimizer with a learning rate of $2e-4$. The models are tested on the tasks introduced in the previous part. And more parameters used in generation are listed in Table \ref{param}.

\textbf{Experiments parameters:}

\begin{table}[htbp]
	\centering
	\caption{Parameters in Validate}
	\label{param}
	\begin{tabular}{c|c|c|c}  
		\hline 
		temperature & top k & top p & length penalty \\ 
		\hline
		1& 50 & 1.0 & 1.0 \\ 
		\hline
	\end{tabular}
\end{table}

\subsection{Details in teacher-student loop in Figure \ref{task_evol} (d)}

\begin{figure*}[ht]
	\includegraphics[width=0.95\textwidth]{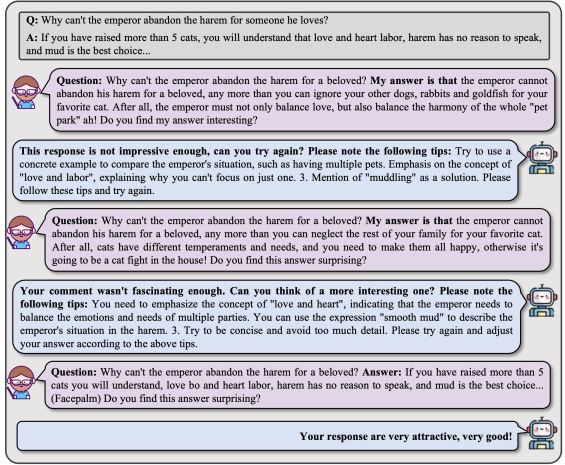}
	\caption{An example of \textit{The teacher-student prompt loop} in Section \ref{sec-diet}}
	\label{exm-tsloop}
\end{figure*}

Through the "teacher-student" interaction mechanism between two agents, the key characteristics of human thinking have been successfully simulated. The trial-and-error mechanism allows for the exploration of different solution paths; the reflection phase promotes an in-depth analysis of the root causes of errors; and the backtracking function enables the dynamic correction of the thinking path. This cognitive architecture makes the reasoning process of AI closer to the real-world problem-solving patterns of human experts.
We utilize the data from the teacher agent to train the model with judgment ability. And an example of this loop is shown in Figure \ref{exm-tsloop}.

% \vspace{-0.5cm}
% \vspace{-1.0cm}

\subsection{Details of AIE}
\label{apd-DIET}

The detailed process of AIE is shown in Algorithm \ref{alg-aaie}. The generator will inject more information into the conversation, and the imaginator will explore the boundaries of the knowledge graph behind the conversation. Finally, the criterion serves as a form of reject sampling for the augmented instructions.

\begin{figure}[h]
	\begin{center}
		\includegraphics[width=0.95\textwidth]{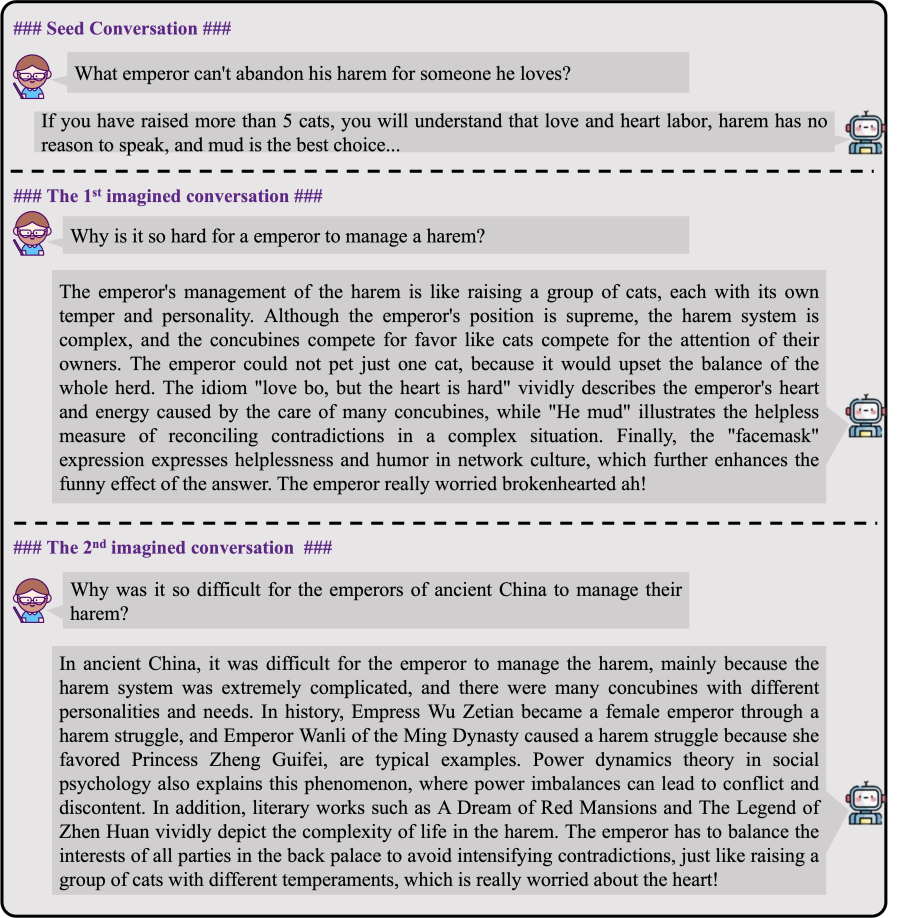}
	\end{center}
	\caption{An example of the imagined process in AIE. It stops according to Criterion-1 in Figure \ref {img-aaie}(b).}
	\label{imagined_conv1}
\end{figure}

\begin{figure}[h]
	\begin{center}
		\includegraphics[width=0.95\textwidth]{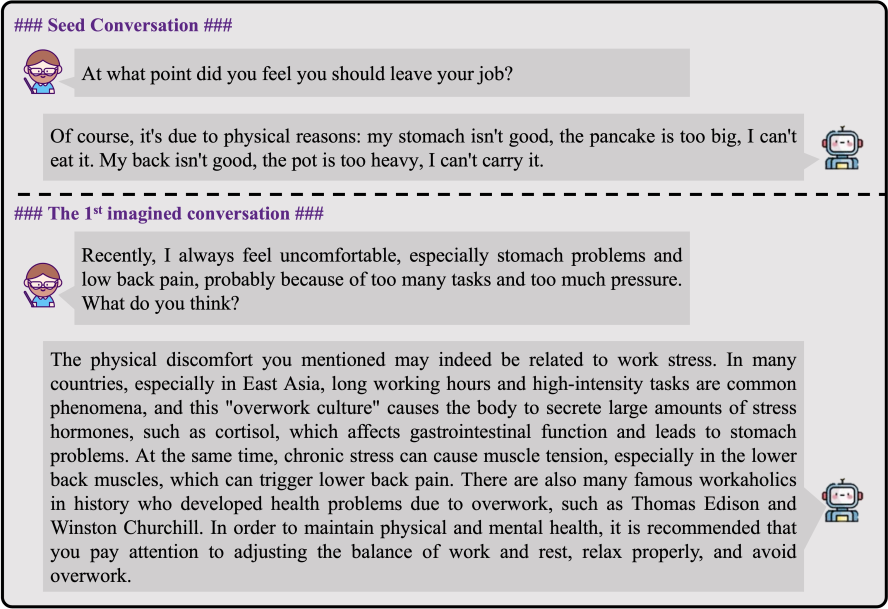}
	\end{center}
	\caption{An example of the imagined process in AIE is presented here. It stops according to Criterion-2 in Figure \ref{img-aaie}(b), i.e., the core idea or joke of the imagined conversation has shifted from the seed conversation.}
	\label{imagined_conv2}
\end{figure}

An example of the generation process is shown in Figure \ref{exm-aaie-a}. Before the evolution, the simple instruction will lead to very simple answers and understanding from the model. While after the evolution, the instruction can become more concrete and the answer will be more informative. And an example of the rule evolution is shown in Figure \ref{exm-aaie-rule}. Additionally, an example of rule evolution is shown in Figure \ref{exm-aaie-rule}. In addition, examples of the imagined conversations are shown in Figure \ref{imagined_conv1} and Figure \ref{imagined_conv2}.

\subsection{Details in GESIT}
\label{apd-GESIT}

GESIT develops an online DPO training frame with rationales to reinforce the reasoning ability. The example of the template is shown in Figure \ref{exm-aaie-rule}. And the training frame is shown in Algorithm \ref{gest}.

\begin{figure*}[ht]
	\includegraphics[width=0.95\textwidth]{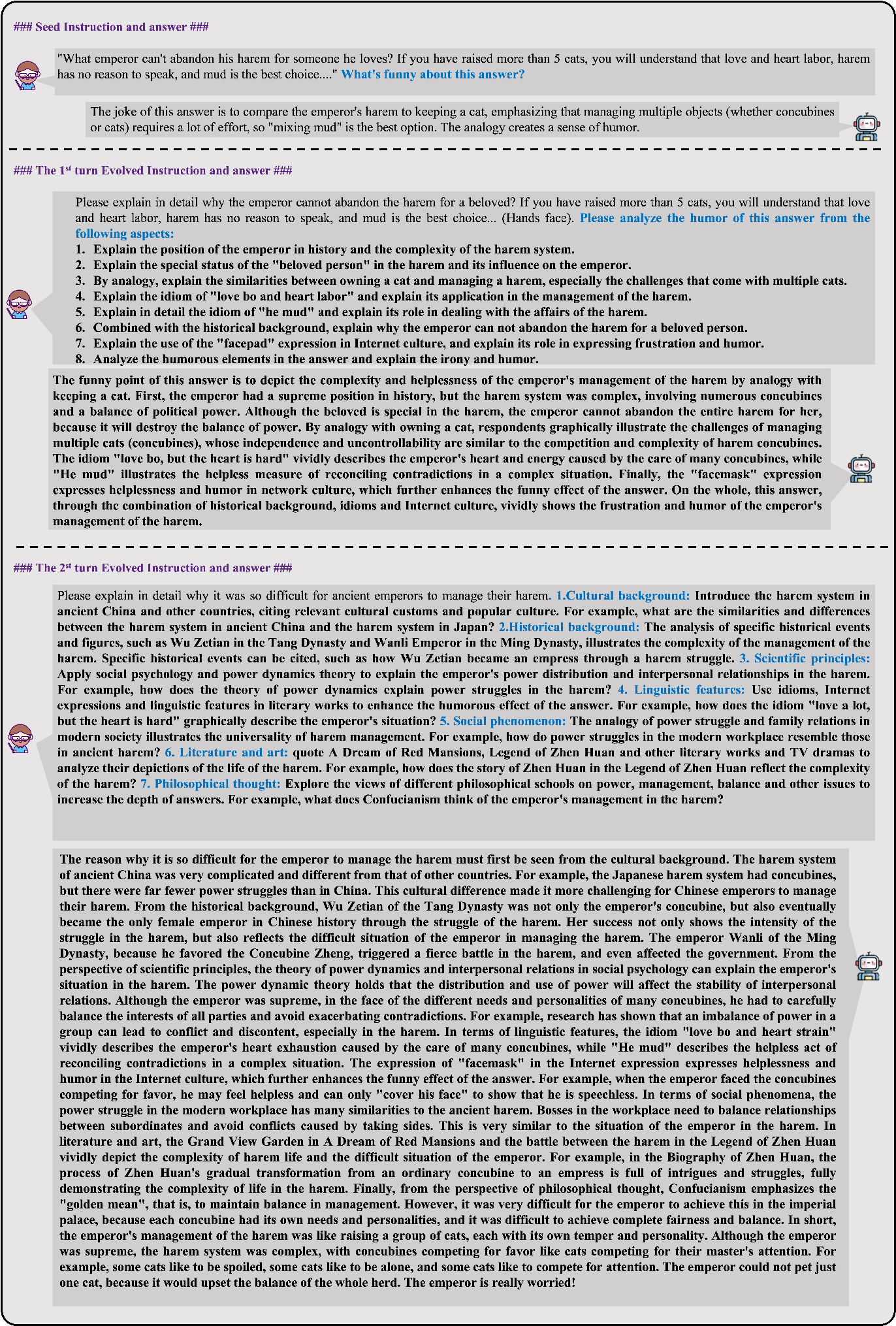}
	\caption{An example of the instruction evolution process in AIE. It stops according to Criterion-1 in Figure \ref {img-aaie}(b).}
	\label{exm-aaie-a}
\end{figure*}

\begin{figure*}[ht]
	\includegraphics[width=0.95\textwidth]{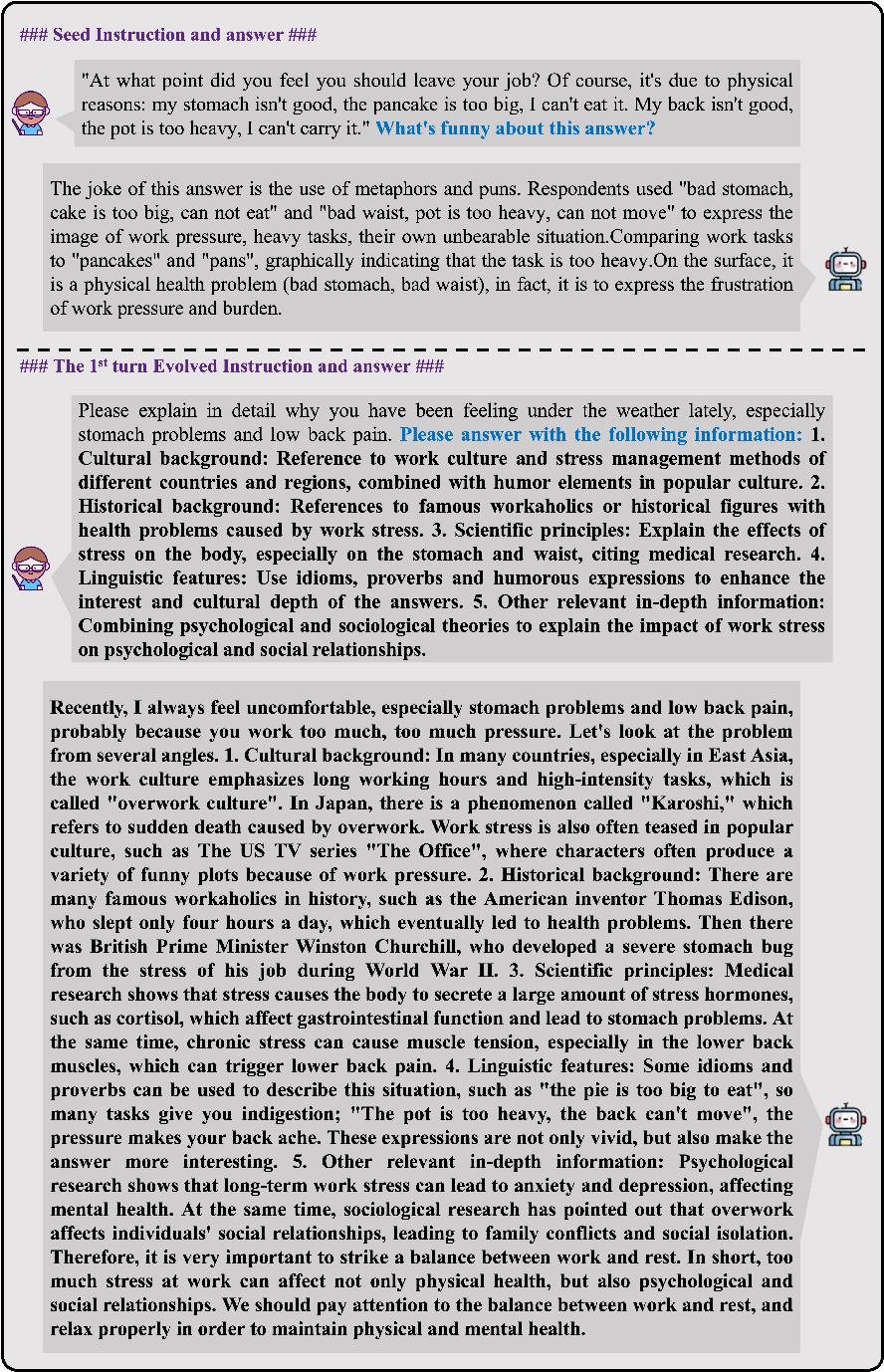}
	\caption{An example of the instruction evolution process in AIE. It stops according to Criterion-2 in Figure \ref {img-aaie}(b).}
	\label{exm-aaie-a}
\end{figure*}

\begin{figure*}[ht]
	\includegraphics[width=0.95\textwidth]{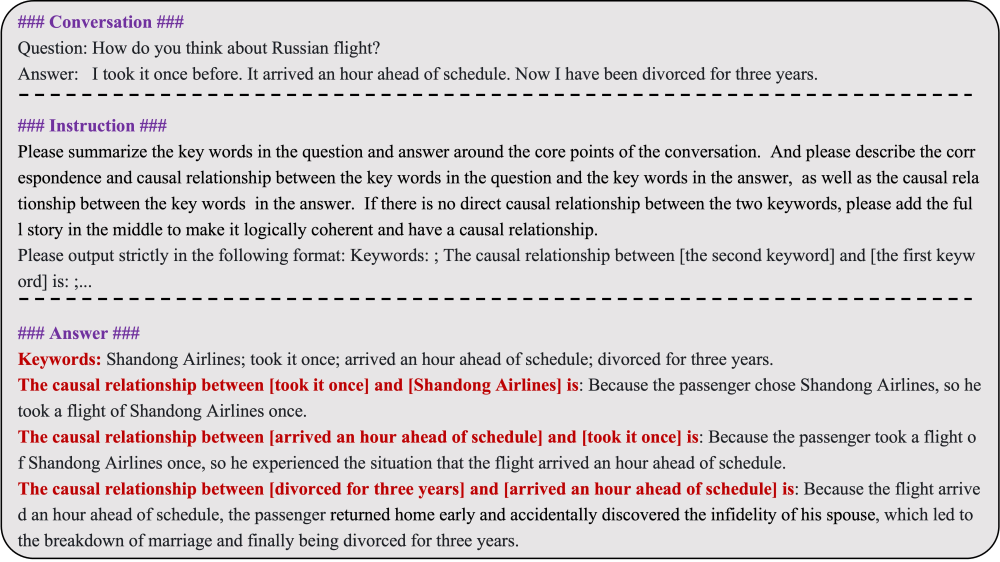}
	\caption{A showcase of the rationale training in GESIT.}
	\label{exm-rationale}
\end{figure*}

\subsection{Generation Showcase}
\label{showcase}

The reasoning algorithm is presented in Algorithm \ref{inf}. We pre-define some generation instructions, such as "Please generate a witty response to the question." We randomly select $n$ instructions, and the model generates $n$ different answers based on these instructions. The model's judgment ability assists in selecting the most humorous answer. The process and results indicate that the model's performance will not be restricted by a specific template.

\textbf{Successful showcase.}
We randomly select posts from the internet and expect the model to provide the most humorous answer. Some examples are presented in Figure \ref{ch-show} and Figure \ref{en-show}. In summary, LoL's replies are usually more concise and direct. It is good at using rhetorical devices such as puns, onomatopoeia, repetition, and rhyme, and combining humor or irony with specific situations, which makes its replies more like "genius replies". In contrast, the responses of GPT-4o and CloT, while sometimes humorous, may be more verbose or lack an immediate punch.

\begin{figure}[ht]
		\begin{minipage}{0.32\linewidth}
				%		\vspace{1pt}
				%这个图片路径替换成你的图片路径即可使用
				\centerline{\includegraphics[width=\textwidth]{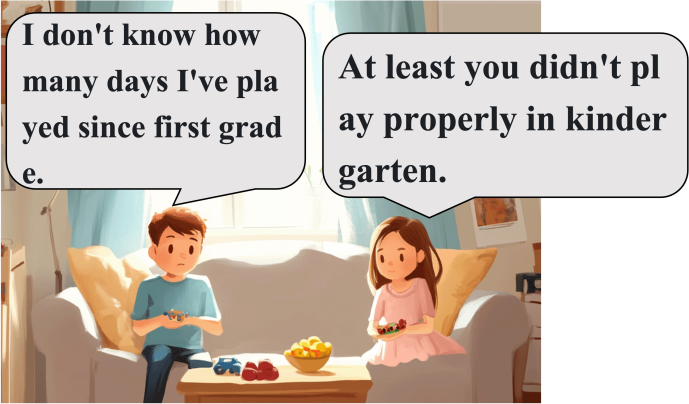}}
				% 加入对这列的图片说明
				\centerline{(a)}
			\end{minipage}
	\begin{minipage}{0.32\linewidth}
		%		\vspace{1pt}
		\centerline{\includegraphics[width=\textwidth]{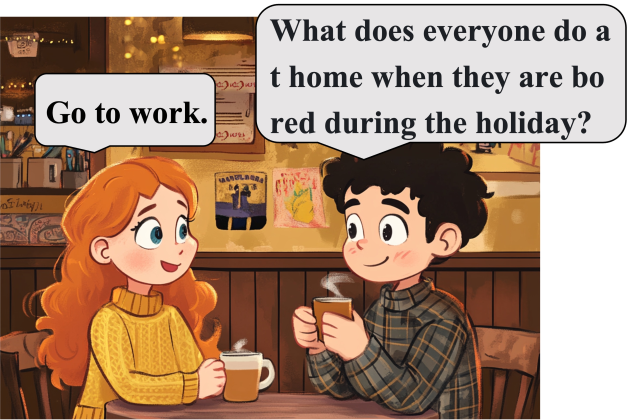}}
		\centerline{(b)}
	\end{minipage}
	\begin{minipage}{0.32\linewidth}
		%		\vspace{1pt}
		\centerline{\includegraphics[width=\textwidth]{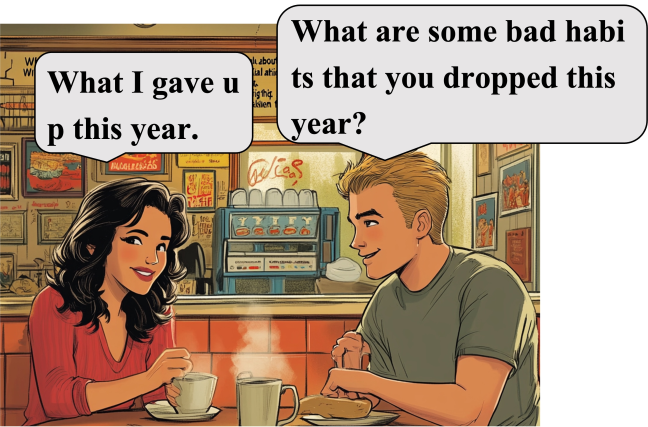}}
		\centerline{(c)}
	\end{minipage}
	\caption{Three showcase of the generation of LoL in English.} 
	\label{showcase-mid-en}
\end{figure}

\begin{figure}[ht]
		\begin{minipage}{0.32\linewidth}
				%		\vspace{1pt}
				%这个图片路径替换成你的图片路径即可使用
				\centerline{\includegraphics[width=\textwidth]{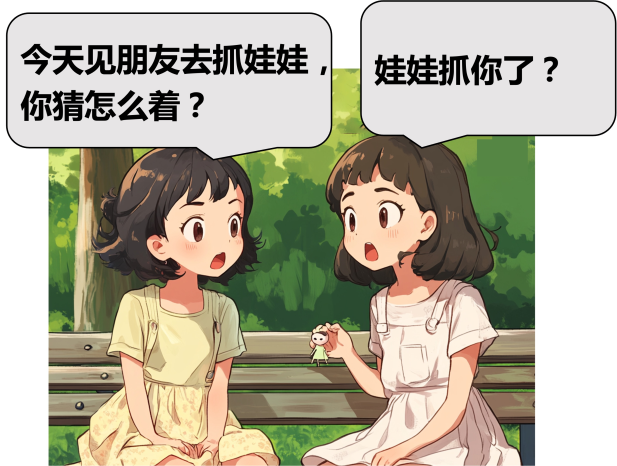}}
				% 加入对这列的图片说明
				\centerline{(a)}
			\end{minipage}
	\begin{minipage}{0.32\linewidth}
		%		\vspace{1pt}
		\centerline{\includegraphics[width=\textwidth]{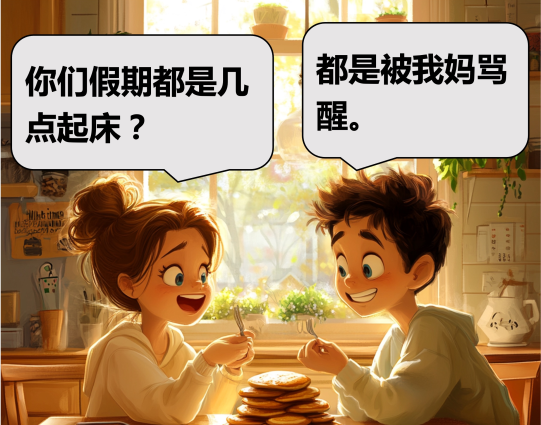}}
		\centerline{(b)}
	\end{minipage}
	\begin{minipage}{0.32\linewidth}
		%		\vspace{1pt}
		\centerline{\includegraphics[width=\textwidth]{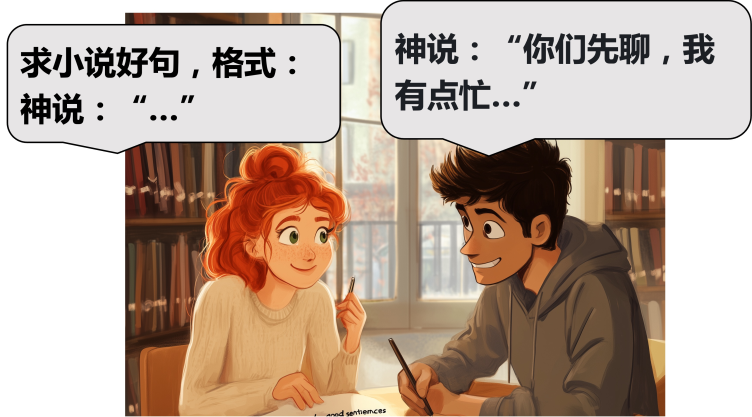}}
		\centerline{(c)}
	\end{minipage}
	\caption{Three showcase of the generation of LoL in Chinese.} 
	\label{showcase-mid-ch}
\end{figure}

\begin{figure}[H]
	\begin{center}
		\includegraphics[width=0.95\textwidth]{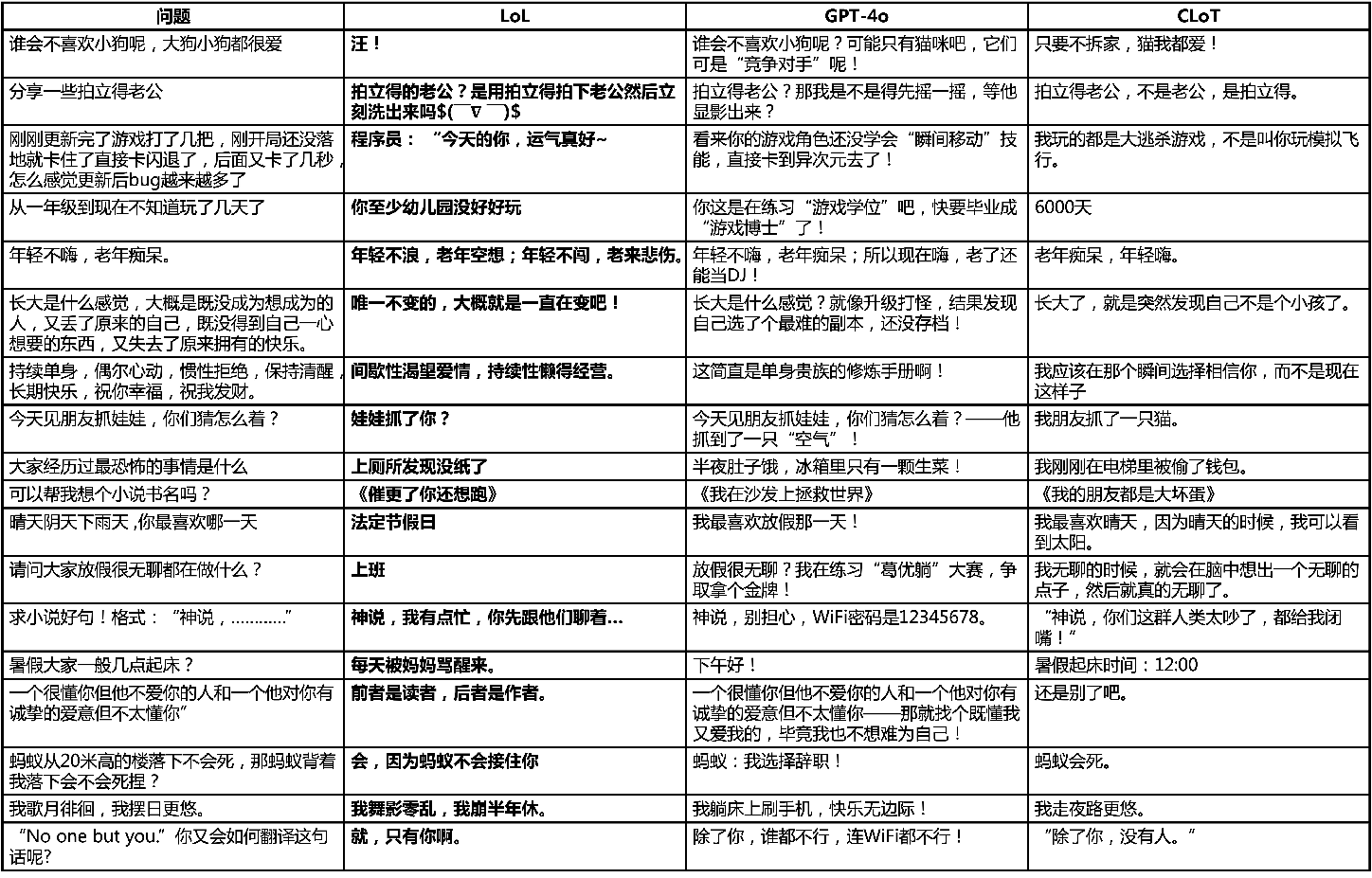}
	\end{center}
	\caption{Chinese Showcases.}
	\label{ch-show}
\end{figure}

\begin{figure}[ht]
	\begin{center}
		\includegraphics[width=0.95\textwidth]{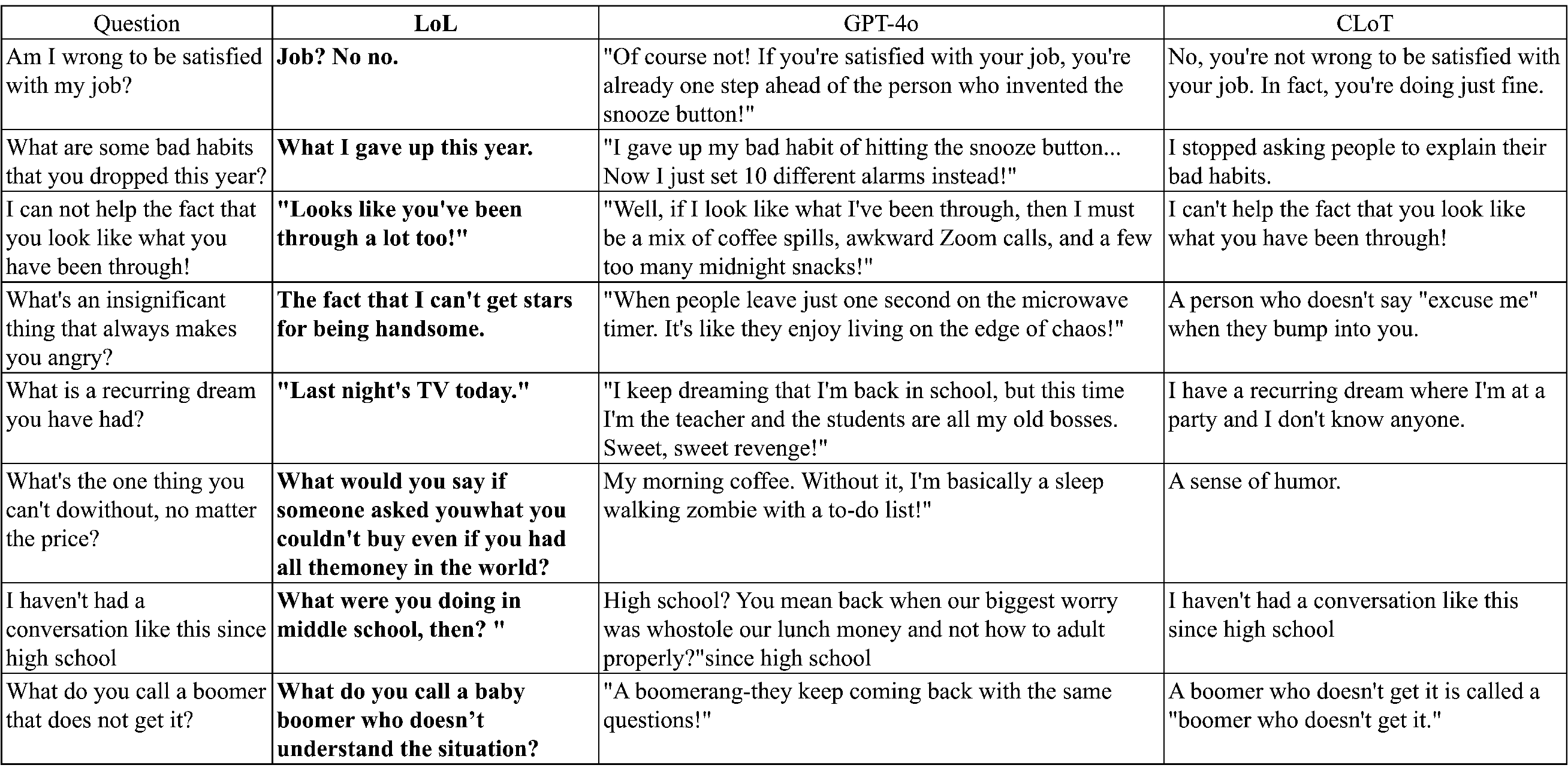}
	\end{center}
	\caption{English Showcases.}
	\label{en-show}
\end{figure}

To better demonstrate an example of the generation capability, we input the conversation into GPT-4o and let it imagine the details of the conversation, such as the scene, the characters, and especially their emotions. Finally, we input the results into Midjourney to generate an image based on the detailed description from GPT-4o. The examples are shown in Figure \ref{showcase-mid-en} and Figure \ref{showcase-mid-ch}.

% \vspace{-2.0cm} % 添加一些垂直间距

\textbf{Failure showcase and Analysis.}
The level of creativity is uneven, and the creativity demonstrated in the samples of the training dataset varies significantly.
For example, in Case 1, LoL's response is an internet meme, while the better-performing one uses a Chinese proverb for comparison.
In terms of creativity, the latter has an edge. In Case 2, a shorter and antithetical answer is more ingenious.
So, There are no failed examples and it's just that according to some people's preferences, they are not humorous.

\begin{table}[htbp]
	\centering
	\caption{Failure case based on user votes}
	\label{param}
	\begin{tabular}{p{3cm}|p{3cm}|p{6cm}}  
		\hline 
		Question & failure case & better than LoL (from human) \\ 
		\hline
		What's your relationship like now?& Had a child. & We are like the weather forecast, sometimes sunny, sometimes rainy, \textcolor{red}{forecasts can never keep up with changes.} \\ 
		\hline
		Why does knowing too much make one an orphan of the world? & Because smart people always find it difficult to find companions, either to become rivals, or to be regarded as rivals.& Because knowing too much makes the world too small.\\
		\hline
	\end{tabular}
\end{table}

\subsection{The Details of User Study}

We conduct a human evaluation to validate LoL's performance in humor generation.
We select the first 200 samples from the validation subset of the Ruozhiba dataset\footnote{https://github.com/Leymore/ruozhiba/tree/main?tab=readme-ov-file} and use the aforementioned method to transform the queries into question-answer pairs. Then, four large language models (LLMs) generate responses to each question, which act as four options.
Subsequently, we conduct a user-preference study to directly evaluate the creativity of the LLMs. We present a question and several corresponding replies and ask users to choose the most creative and humorous response. We select four advanced LLMs to generate responses for a total of 200 questions, and the four responses from the four different LLMs are randomly arranged among the options.
We conduct an extensive survey through an online survey platform\footnote{https://www.wjx.cn/}, ultimately collecting 15 valid questionnaires with 3000 votes. From these collected questionnaires, we calculate the proportion of times each LLM is selected for each question. Finally, we sum up the total number of times each LLM is chosen across all validation samples, as shown in Figure \ref{fig5}(c). The ratio of this sum to the total number of selections among all LLMs represents the user preference for each LLM. Additionally, we calculate the win - rate based on the question dimension, as depicted in Figure \ref{fig5}(b).

\begin{figure}[H]
	\begin{center}
		\includegraphics[width=0.95\textwidth]{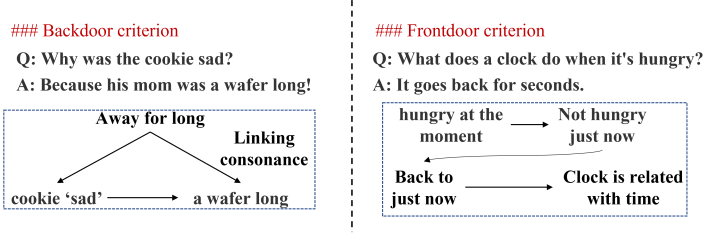}
	\end{center}
	\caption{Backdoor and Frontdoor criterion examples of humor generation.}
	\label{exp_back_front}
\end{figure}

\begin{algorithm}[H]
\caption{Algorithm of Automatic Instruction Expansion}
\begin{algorithmic}[1]
\label{alg-aaie}
\REQUIRE Dialogue dataset $D$, initial instruction $I_0$, evolution rules $R_0$
% \ENSURE Enhanced instruction set $I_{final}$
\STATE Sample initial dialogue $d_0 = (q, a)$ from $D$
\STATE Initialize evolution round counter $t \gets 0$
\STATE Set current dialogue $d_{cur} \gets d_0$
\STATE Set current instruction $I_{cur} \gets I_0$
\STATE Set current rule $R_{cur} \gets R_0$
\WHILE{$t < 3$}
    \STATE // Instruction Evolution Phase
    \STATE $I_{new} \gets \text{Generator}(d_{cur}, I_{cur}, R_{cur})$  \hfill \mycomment{ // as shown in Figure \ref{img-aaie}(a)}
    \STATE $d_{imag} \gets \text{Imaginator}(I_{new}, a)$ \hfill \mycomment{ // as shown in Figure \ref{img-aaie}(a)}
    \STATE // Analysis Phase \hfill \mycomment{ // as shown in Figure \ref{img-aaie}(b)}
    \IF{$ \text{Criterion-2}(d_{imag})$} 
        \STATE \textbf{break}
    \ENDIF
    \IF{$\neg \text{Criterion-1}(I_{new}, I_{cur})$}
        \STATE $d_{cur} \gets d_{imag}$
        \STATE $R_{cur} \gets \text{Analyst}(R_0, I_{cur})$
        \STATE $t \gets t + 1$
    \ELSE
        \STATE $I_{cur} \gets I_{new}$
        \STATE \textbf{break}
    \ENDIF
\ENDWHILE
\STATE Store $I_{cur}, d_{cur}$
\end{algorithmic}
\end{algorithm}

\begin{algorithm}[ht]
\caption{Algorithm of GESIT}
\label{gest}
\renewcommand{\algorithmicrequire}{\textbf{Input:}}
\renewcommand{\algorithmicensure}{\textbf{Output:}}
\begin{algorithmic}[1]
		\REQUIRE Policy model $\pi$, Discriminator $\pi^*$, Preference dataset $D_0=\{( q_i, a_i^+, a_i^-) \}$, Expert $E$.  %%input
		\ENSURE Finetuned policy model $\pi$   %%output   
		\STATE $\widetilde D \leftarrow \{\}$, $D_k \leftarrow D_0, k\leftarrow0$,
		\FOR {each $t \in N$}
    		\STATE sample $(q, a^+, a^-) \in D_k$ 
    		\STATE $\pi \leftarrow \pi + \nabla L_{DPO}(\pi, (\widetilde q, a^+, a^-))$  \hfill // DPO training on origin dataset.
    		\IF {$t = T$}
        		\STATE sample $\widetilde q \in D_k$  \hfill // Randomly sample questions form origin dataset.
        		\STATE $\widetilde a_1, \widetilde a_2 \leftarrow \ \ \pi(\widetilde q), \pi(\widetilde q)$  \hfill // $\pi$ outputs new responses to sampled questions.
        		\STATE $\widetilde a^+, \widetilde a^-\leftarrow \ \ \pi^*(\widetilde q, \widetilde a_1, \widetilde a_2)$ \hfill // $\pi^*$ judges both responses as positive and negative samples.
        		\STATE $\widetilde r^+, \widetilde r^-\leftarrow \ \ E(\widetilde q, \widetilde a^+), E(q, \widetilde a^-)$  \hfill // $E$ outputs rationales for responses.
        		\STATE $\widetilde D \leftarrow \{ \widetilde q, \widetilde a^+, \widetilde a^-, \widetilde r^+, \widetilde r^-\}$ 
        		\STATE $D_{k+1} \leftarrow D_k \cup \widetilde D$  \hfill // New data are added into dataset to train.
    		\ENDIF
		\ENDFOR
\end{algorithmic}
\end{algorithm}

\begin{algorithm}[h]
	\caption{Inference Step of LoL}\label{inf}
	\renewcommand{\algorithmicrequire}{\textbf{Input:}}
	\renewcommand{\algorithmicensure}{\textbf{Output:}}
	\begin{algorithmic}
		\REQUIRE Questions $Q$, LoL-trained LLM $\pi'$, response number $n$, different prompts $P$.
		\ENSURE The most funny response $A_{best}$
		\STATE Select $n$ Prompts $P_{sub}$ from $P$
		\STATE $\{A_i\}_{i=0}^n \leftarrow \pi'(Q, P_{sub})$  \hfill // $\pi'$ output candidate responses through $P_{sub}$.
		\STATE $A_{best} \leftarrow \pi'(Q, \{A_i\}_{i=0}^n)$  \hfill // $\pi'$ as judgement model to select the most funny answer.
		% \UNTIL{$A_{best}$}
	\end{algorithmic}
\end{algorithm}

\begin{figure}[H]
	\begin{center}
		\includegraphics[width=0.95\textwidth]{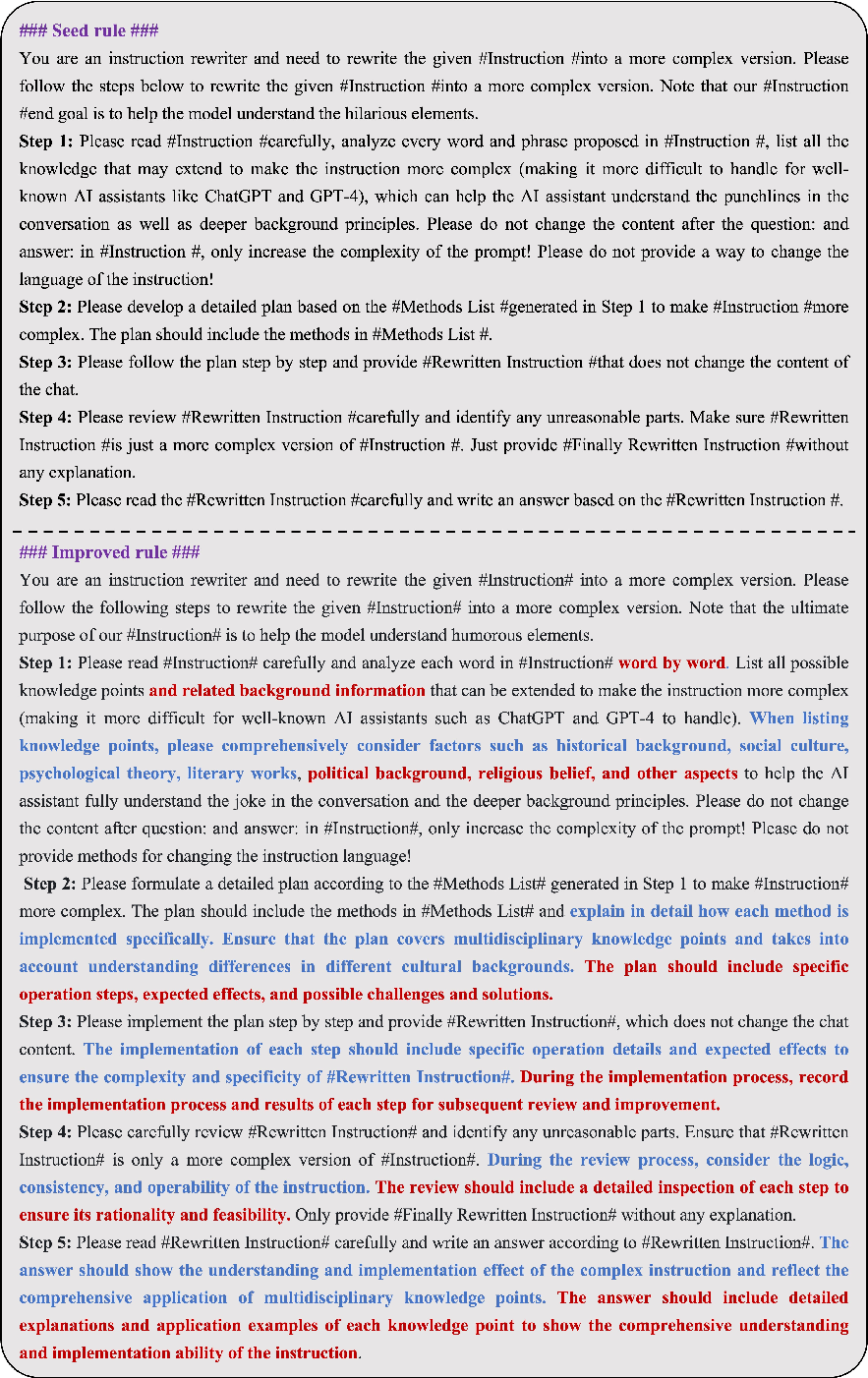}
	\end{center}
	\caption{The seed rule and an example of evolved rule in AIE mentioned in Section \ref{sec-diet}. What is marked in blue is the result after the first rule improvement (line 16 in Algorithm \ref{alg-aaie}), while the text in red is the result after the second rule improvement.}
	\label{exm-aaie-rule}
\end{figure}

%%%%%%%%%%%%%%%%%%%%%%%%%%%%%%%%%%%%%%%%%%%%%%%%%%%%%%%%%%%%%%%%%%%%%%%%%%%%%%%
%%%%%%%%%%%%%%%%%%%%%%%%%%%%%%%%%%%%%%%%%%%%%%%%%%%%%%%%%%%%%%%%%%%%%%%%%%%%%%%

\end{document}

%% file: math_commands.tex
\usepackage{amsmath,amsfonts,bm}

\def\eqref#1{equation~\ref{#1}}
\def\1{\bm{1}}

\DeclareMathAlphabet{\mathsfit}{\encodingdefault}{\sfdefault}{m}{sl}
\SetMathAlphabet{\mathsfit}{bold}{\encodingdefault}{\sfdefault}{bx}{n}

%% file: iclr2025_conference.bbl
\begin{thebibliography}{75}
\providecommand{\natexlab}[1]{#1}
\providecommand{\url}[1]{\texttt{#1}}
\expandafter\ifx\csname urlstyle\endcsname\relax
  \providecommand{\doi}[1]{doi: #1}\else
  \providecommand{\doi}{doi: \begingroup \urlstyle{rm}\Url}\fi

\bibitem[Amin \& Burghardt(2020)Amin and Burghardt]{amin2020survey}
Miriam Amin and Manuel Burghardt.
\newblock A survey on approaches to computational humor generation.
\newblock In \emph{Proceedings of the 4th Joint SIGHUM Workshop on
  Computational Linguistics for Cultural Heritage, Social Sciences, Humanities
  and Literature}, pp.\  29--41, 2020.

\bibitem[Bai et~al.(2023)Bai, Bai, Yang, Wang, Tan, Wang, Lin, Zhou, and
  Zhou]{bai2023qwen}
Jinze Bai, Shuai Bai, Shusheng Yang, Shijie Wang, Sinan Tan, Peng Wang, Junyang
  Lin, Chang Zhou, and Jingren Zhou.
\newblock Qwen-vl: A versatile vision-language model for understanding,
  localization, text reading, and beyond.
\newblock 2023.

\bibitem[Bhavya et~al.(2023)Bhavya, Xiong, and Zhai]{bhavya2023cam}
Bhavya Bhavya, Jinjun Xiong, and Chengxiang Zhai.
\newblock Cam: A large language model-based creative analogy mining framework.
\newblock In \emph{Proceedings of the ACM Web Conference 2023}, pp.\
  3903--3914, 2023.

\bibitem[Binsted et~al.(2006)Binsted, Nijholt, Stock, Strapparava, Ritchie,
  Manurung, Pain, Waller, and O'Mara]{binsted2006computational}
Kim Binsted, Anton Nijholt, Oliviero Stock, Carlo Strapparava, G~Ritchie,
  R~Manurung, H~Pain, Annalu Waller, and D~O'Mara.
\newblock Computational humor.
\newblock \emph{IEEE intelligent systems}, 21\penalty0 (2):\penalty0 59--69,
  2006.

\bibitem[Chakrabarty et~al.(2022)Chakrabarty, Padmakumar, and
  He]{chakrabarty2022help}
Tuhin Chakrabarty, Vishakh Padmakumar, and He~He.
\newblock Help me write a poem: Instruction tuning as a vehicle for
  collaborative poetry writing.
\newblock \emph{arXiv preprint arXiv:2210.13669}, 2022.

\bibitem[Chaudhary et~al.(2021)Chaudhary, Goel, and
  Mamidi]{chaudhary2021towards}
Tanishq Chaudhary, Mayank Goel, and Radhika Mamidi.
\newblock Towards conversational humor analysis and design.
\newblock \emph{arXiv preprint arXiv:2103.00536}, 2021.

\bibitem[Chen \& Zhang(2022)Chen and Zhang]{chen2022integrating}
Chengxin Chen and Pengyuan Zhang.
\newblock Integrating cross-modal interactions via latent representation shift
  for multi-modal humor detection.
\newblock In \emph{Proceedings of the 3rd International on Multimodal Sentiment
  Analysis Workshop and Challenge}, pp.\  23--28, 2022.

\bibitem[Chen et~al.(2023)Chen, Zhu, Shen, Li, Liu, Zhang, Krishnamoorthi,
  Chandra, Xiong, and Elhoseiny]{chen2023minigpt}
Jun Chen, Deyao Zhu, Xiaoqian Shen, Xiang Li, Zechun Liu, Pengchuan Zhang,
  Raghuraman Krishnamoorthi, Vikas Chandra, Yunyang Xiong, and Mohamed
  Elhoseiny.
\newblock Minigpt-v2: large language model as a unified interface for
  vision-language multi-task learning.
\newblock \emph{arXiv preprint arXiv:2310.09478}, 2023.

\bibitem[Chiang et~al.(2023)Chiang, Li, Lin, Sheng, Wu, Zhang, Zheng, Zhuang,
  Zhuang, Gonzalez, et~al.]{chiang2023vicuna}
Wei-Lin Chiang, Zhuohan Li, Zi~Lin, Ying Sheng, Zhanghao Wu, Hao Zhang, Lianmin
  Zheng, Siyuan Zhuang, Yonghao Zhuang, Joseph~E Gonzalez, et~al.
\newblock Vicuna: An open-source chatbot impressing gpt-4 with 90\%* chatgpt
  quality.
\newblock \emph{See https://vicuna. lmsys. org (accessed 14 April 2023)},
  2\penalty0 (3):\penalty0 6, 2023.

\bibitem[Dang et~al.(2023)Dang, Goller, Lehmann, and Buschek]{dang2023choice}
Hai Dang, Sven Goller, Florian Lehmann, and Daniel Buschek.
\newblock Choice over control: How users write with large language models using
  diegetic and non-diegetic prompting.
\newblock In \emph{Proceedings of the 2023 CHI Conference on Human Factors in
  Computing Systems}, pp.\  1--17, 2023.

\bibitem[Dong et~al.(2023)Dong, Han, Peng, Qi, Ge, Yang, Zhao, Sun, Zhou, Wei,
  et~al.]{dong2023dreamllm}
Runpei Dong, Chunrui Han, Yuang Peng, Zekun Qi, Zheng Ge, Jinrong Yang, Liang
  Zhao, Jianjian Sun, Hongyu Zhou, Haoran Wei, et~al.
\newblock Dreamllm: Synergistic multimodal comprehension and creation.
\newblock \emph{arXiv preprint arXiv:2309.11499}, 2023.

\bibitem[Driess et~al.(2023)Driess, Xia, Sajjadi, Lynch, Chowdhery, Ichter,
  Wahid, Tompson, Vuong, Yu, et~al.]{driess2023palm}
Danny Driess, Fei Xia, Mehdi~SM Sajjadi, Corey Lynch, Aakanksha Chowdhery,
  Brian Ichter, Ayzaan Wahid, Jonathan Tompson, Quan Vuong, Tianhe Yu, et~al.
\newblock Palm-e: An embodied multimodal language model.
\newblock \emph{arXiv preprint arXiv:2303.03378}, 2023.

\bibitem[Evans et~al.(2019)Evans, Slaughter, Ellis, and Rivin]{evans2019gender}
Jonathan~B Evans, Jerel~E Slaughter, Aleksander~PJ Ellis, and Jessi~M Rivin.
\newblock Gender and the evaluation of humor at work.
\newblock \emph{Journal of Applied Psychology}, 104\penalty0 (8):\penalty0
  1077, 2019.

\bibitem[Fu et~al.(2022)Fu, Peng, Sabharwal, Clark, and Khot]{fu2022complexity}
Yao Fu, Hao Peng, Ashish Sabharwal, Peter Clark, and Tushar Khot.
\newblock Complexity-based prompting for multi-step reasoning.
\newblock In \emph{The Eleventh International Conference on Learning
  Representations}, 2022.

\bibitem[Hope et~al.(2022)Hope, Tamari, Hershcovich, Kang, Chan, Kittur, and
  Shahaf]{hope2022scaling}
Tom Hope, Ronen Tamari, Daniel Hershcovich, Hyeonsu~B Kang, Joel Chan, Aniket
  Kittur, and Dafna Shahaf.
\newblock Scaling creative inspiration with fine-grained functional aspects of
  ideas.
\newblock In \emph{Proceedings of the 2022 CHI Conference on Human Factors in
  Computing Systems}, pp.\  1--15, 2022.

\bibitem[Hossain et~al.(2020{\natexlab{a}})Hossain, Krumm, Gamon, and
  Kautz]{hossain-etal-2020-semeval}
Nabil Hossain, John Krumm, Michael Gamon, and Henry Kautz.
\newblock {S}em{E}val-2020 task 7: Assessing humor in edited news headlines.
\newblock In Aurelie Herbelot, Xiaodan Zhu, Alexis Palmer, Nathan Schneider,
  Jonathan May, and Ekaterina Shutova (eds.), \emph{Proceedings of the
  Fourteenth Workshop on Semantic Evaluation}, pp.\  746--758, Barcelona
  (online), December 2020{\natexlab{a}}. International Committee for
  Computational Linguistics.
\newblock \doi{10.18653/v1/2020.semeval-1.98}.
\newblock URL \url{https://aclanthology.org/2020.semeval-1.98}.

\bibitem[Hossain et~al.(2020{\natexlab{b}})Hossain, Krumm, Sajed, and
  Kautz]{hossain2020stimulating}
Nabil Hossain, John Krumm, Tanvir Sajed, and Henry Kautz.
\newblock Stimulating creativity with funlines: A case study of humor
  generation in headlines.
\newblock \emph{arXiv preprint arXiv:2002.02031}, 2020{\natexlab{b}}.

\bibitem[Hu et~al.(2021)Hu, Shen, Wallis, Allen-Zhu, Li, Wang, Wang, and
  Chen]{hu2021lora}
Edward~J Hu, Yelong Shen, Phillip Wallis, Zeyuan Allen-Zhu, Yuanzhi Li, Shean
  Wang, Lu~Wang, and Weizhu Chen.
\newblock Lora: Low-rank adaptation of large language models.
\newblock \emph{arXiv preprint arXiv:2106.09685}, 2021.

\bibitem[Huang et~al.(2023)Huang, Liang, Zhang, Yang, and Lin]{huang2023fast}
Zhongzhan Huang, Senwei Liang, Hong Zhang, Haizhao Yang, and Liang Lin.
\newblock On fast simulation of dynamical system with neural vector enhanced
  numerical solver.
\newblock \emph{Scientific reports}, 13\penalty0 (1):\penalty0 15254, 2023.

\bibitem[Hwang \& Shwartz(2023)Hwang and Shwartz]{hwang2023memecap}
EunJeong Hwang and Vered Shwartz.
\newblock Memecap: A dataset for captioning and interpreting memes.
\newblock \emph{arXiv preprint arXiv:2305.13703}, 2023.

\bibitem[Jiang et~al.(2023)Jiang, Wang, Zeng, Zhong, Li, Mi, Shang, Jiang, Liu,
  and Wang]{jiang2023followbench}
Yuxin Jiang, Yufei Wang, Xingshan Zeng, Wanjun Zhong, Liangyou Li, Fei Mi,
  Lifeng Shang, Xin Jiang, Qun Liu, and Wei Wang.
\newblock Followbench: A multi-level fine-grained constraints following
  benchmark for large language models.
\newblock \emph{arXiv preprint arXiv:2310.20410}, 2023.

\bibitem[Kang et~al.(2022)Kang, Qian, Hope, Shahaf, Chan, and
  Kittur]{kang2022augmenting}
Hyeonsu~B Kang, Xin Qian, Tom Hope, Dafna Shahaf, Joel Chan, and Aniket Kittur.
\newblock Augmenting scientific creativity with an analogical search engine.
\newblock \emph{ACM Transactions on Computer-Human Interaction}, 29\penalty0
  (6):\penalty0 1--36, 2022.

\bibitem[Kumar et~al.(2022)Kumar, Walia, and Sharma]{kumar2022deephumor}
Vijay Kumar, Ranjeet Walia, and Shivam Sharma.
\newblock Deephumor: a novel deep learning framework for humor detection.
\newblock \emph{Multimedia Tools and Applications}, 81\penalty0 (12):\penalty0
  16797--16812, 2022.

\bibitem[Lai et~al.(2024)Lai, Tian, Chen, Yang, Peng, and Jia]{lai2024step}
Xin Lai, Zhuotao Tian, Yukang Chen, Senqiao Yang, Xiangru Peng, and Jiaya Jia.
\newblock Step-dpo: Step-wise preference optimization for long-chain reasoning
  of llms.
\newblock \emph{arXiv preprint arXiv:2406.18629}, 2024.

\bibitem[Li et~al.(2018)Li, Zhao, Hu, Li, Liu, and Du]{li2018analogical}
Shen Li, Zhe Zhao, Renfen Hu, Wensi Li, Tao Liu, and Xiaoyong Du.
\newblock Analogical reasoning on chinese morphological and semantic relations.
\newblock \emph{arXiv preprint arXiv:1805.06504}, 2018.

\bibitem[Li et~al.(2023)Li, Yu, Zhou, Schick, Levy, Zettlemoyer, Weston, and
  Lewis]{li2023self}
Xian Li, Ping Yu, Chunting Zhou, Timo Schick, Omer Levy, Luke Zettlemoyer,
  Jason Weston, and Mike Lewis.
\newblock Self-alignment with instruction backtranslation.
\newblock \emph{arXiv preprint arXiv:2308.06259}, 2023.

\bibitem[Liang et~al.(2024)Liang, Wu, et~al.]{liang2024toa}
Mingfu Liang, Ying Wu, et~al.
\newblock Toa: task-oriented active vqa.
\newblock \emph{Advances in Neural Information Processing Systems}, 36, 2024.

\bibitem[Liang et~al.(2022)Liang, Huang, and Zhang]{liang2022stiffness}
Senwei Liang, Zhongzhan Huang, and Hong Zhang.
\newblock Stiffness-aware neural network for learning hamiltonian systems.
\newblock In \emph{International Conference on Learning Representations}, 2022.

\bibitem[Lightman et~al.(2023)Lightman, Kosaraju, Burda, Edwards, Baker, Lee,
  Leike, Schulman, Sutskever, and Cobbe]{lightman2023let}
Hunter Lightman, Vineet Kosaraju, Yura Burda, Harri Edwards, Bowen Baker, Teddy
  Lee, Jan Leike, John Schulman, Ilya Sutskever, and Karl Cobbe.
\newblock Let's verify step by step.
\newblock \emph{arXiv preprint arXiv:2305.20050}, 2023.

\bibitem[Ling et~al.(2023)Ling, Fang, Li, Mu, Lee, Pourreza, Memisevic, and
  Su]{ling2023unleashing}
Zhan Ling, Yunhao Fang, Xuanlin Li, Tongzhou Mu, Mingu Lee, Reza Pourreza,
  Roland Memisevic, and Hao Su.
\newblock Unleashing the creative mind: Language model as hierarchical policy
  for improved exploration on challenging problem solving.
\newblock 2023.

\bibitem[Liu et~al.(2024)Liu, Li, Li, and Lee]{liu2024improved}
Haotian Liu, Chunyuan Li, Yuheng Li, and Yong~Jae Lee.
\newblock Improved baselines with visual instruction tuning.
\newblock In \emph{Proceedings of the IEEE/CVF Conference on Computer Vision
  and Pattern Recognition}, pp.\  26296--26306, 2024.

\bibitem[Long(2023)]{long2023large}
Jieyi Long.
\newblock Large language model guided tree-of-thought.
\newblock \emph{arXiv preprint arXiv:2305.08291}, 2023.

\bibitem[Luong et~al.(2024)Luong, Zhang, Jie, Sun, Jin, and Li]{luong2024reft}
Trung~Quoc Luong, Xinbo Zhang, Zhanming Jie, Peng Sun, Xiaoran Jin, and Hang
  Li.
\newblock Reft: Reasoning with reinforced fine-tuning.
\newblock \emph{arXiv preprint arXiv:2401.08967}, 2024.

\bibitem[Meaney et~al.(2021)Meaney, Wilson, Chiruzzo, Lopez, and
  Magdy]{meaney2021semeval}
JA~Meaney, Steven Wilson, Luis Chiruzzo, Adam Lopez, and Walid Magdy.
\newblock Semeval 2021 task 7: Hahackathon, detecting and rating humor and
  offense.
\newblock In \emph{Proceedings of the 15th International Workshop on Semantic
  Evaluation (SemEval-2021)}, pp.\  105--119, 2021.

\bibitem[Mirowski et~al.(2023)Mirowski, Mathewson, Pittman, and
  Evans]{mirowski2023co}
Piotr Mirowski, Kory~W Mathewson, Jaylen Pittman, and Richard Evans.
\newblock Co-writing screenplays and theatre scripts with language models:
  Evaluation by industry professionals.
\newblock In \emph{Proceedings of the 2023 CHI Conference on Human Factors in
  Computing Systems}, pp.\  1--34, 2023.

\bibitem[o1~Team(2024)]{skyworkopeno12024}
Skywork o1~Team.
\newblock Skywork-o1 open series.
\newblock \url{https://huggingface.co/Skywork}, November 2024.
\newblock URL \url{https://huggingface.co/Skywork}.

\bibitem[Ofer \& Shahaf(2022)Ofer and Shahaf]{ofer2022cards}
Dan Ofer and Dafna Shahaf.
\newblock Cards against ai: Predicting humor in a fill-in-the-blank party game.
\newblock \emph{arXiv preprint arXiv:2210.13016}, 2022.

\bibitem[Olson et~al.(2021)Olson, Nahas, Chmoulevitch, Cropper, and
  Webb]{olson2021naming}
Jay~A Olson, Johnny Nahas, Denis Chmoulevitch, Simon~J Cropper, and Margaret~E
  Webb.
\newblock Naming unrelated words predicts creativity.
\newblock \emph{Proceedings of the National Academy of Sciences}, 118\penalty0
  (25):\penalty0 e2022340118, 2021.

\bibitem[Park et~al.(2023)Park, Leahey, and Funk]{park2023papers}
Michael Park, Erin Leahey, and Russell~J Funk.
\newblock Papers and patents are becoming less disruptive over time.
\newblock \emph{Nature}, 613\penalty0 (7942):\penalty0 138--144, 2023.

\bibitem[Qin et~al.(2024{\natexlab{a}})Qin, Li, Zou, Liu, Xia, Huang, Ye, Yuan,
  Liu, Li, et~al.]{qin2024o1}
Yiwei Qin, Xuefeng Li, Haoyang Zou, Yixiu Liu, Shijie Xia, Zhen Huang, Yixin
  Ye, Weizhe Yuan, Hector Liu, Yuanzhi Li, et~al.
\newblock O1 replication journey: A strategic progress report--part 1.
\newblock \emph{arXiv preprint arXiv:2410.18982}, 2024{\natexlab{a}}.

\bibitem[Qin et~al.(2024{\natexlab{b}})Qin, Song, Hu, Yao, Cho, Wang, Wu, Liu,
  Liu, and Yu]{qin2024infobench}
Yiwei Qin, Kaiqiang Song, Yebowen Hu, Wenlin Yao, Sangwoo Cho, Xiaoyang Wang,
  Xuansheng Wu, Fei Liu, Pengfei Liu, and Dong Yu.
\newblock Infobench: Evaluating instruction following ability in large language
  models.
\newblock \emph{arXiv preprint arXiv:2401.03601}, 2024{\natexlab{b}}.

\bibitem[Rafailov et~al.(2024)Rafailov, Sharma, Mitchell, Manning, Ermon, and
  Finn]{rafailov2024direct}
Rafael Rafailov, Archit Sharma, Eric Mitchell, Christopher~D Manning, Stefano
  Ermon, and Chelsea Finn.
\newblock Direct preference optimization: Your language model is secretly a
  reward model.
\newblock \emph{Advances in Neural Information Processing Systems}, 36, 2024.

\bibitem[Saparov \& He(2022)Saparov and He]{saparov2022language}
Abulhair Saparov and He~He.
\newblock Language models are greedy reasoners: A systematic formal analysis of
  chain-of-thought.
\newblock \emph{arXiv preprint arXiv:2210.01240}, 2022.

\bibitem[Shahaf et~al.(2015)Shahaf, Horvitz, and Mankoff]{shahaf2015inside}
Dafna Shahaf, Eric Horvitz, and Robert Mankoff.
\newblock Inside jokes: Identifying humorous cartoon captions.
\newblock In \emph{Proceedings of the 21th ACM SIGKDD international conference
  on knowledge discovery and data mining}, pp.\  1065--1074, 2015.

\bibitem[Summers-Stay et~al.(2023)Summers-Stay, Voss, and
  Lukin]{summers2023brainstorm}
Douglas Summers-Stay, Clare~R Voss, and Stephanie~M Lukin.
\newblock Brainstorm, then select: a generative language model improves its
  creativity score.
\newblock In \emph{The AAAI-23 Workshop on Creative AI Across Modalities},
  2023.

\bibitem[Sun et~al.(2024)Sun, Liu, Li, Wang, Dong, Lin, and
  Huang]{sun2024conifer}
Haoran Sun, Lixin Liu, Junjie Li, Fengyu Wang, Baohua Dong, Ran Lin, and Ruohui
  Huang.
\newblock Conifer: Improving complex constrained instruction-following ability
  of large language models.
\newblock \emph{arXiv preprint arXiv:2404.02823}, 2024.

\bibitem[Sun et~al.(2023{\natexlab{a}})Sun, Xu, Tang, Wang, Lin, Gong, Shum,
  and Guo]{sun2023think}
Jiashuo Sun, Chengjin Xu, Lumingyuan Tang, Saizhuo Wang, Chen Lin, Yeyun Gong,
  Heung-Yeung Shum, and Jian Guo.
\newblock Think-on-graph: Deep and responsible reasoning of large language
  model with knowledge graph.
\newblock \emph{arXiv preprint arXiv:2307.07697}, 2023{\natexlab{a}}.

\bibitem[Sun et~al.(2023{\natexlab{b}})Sun, Li, Peng, and Gao]{sun2023inspire}
Yuqian Sun, Xingyu Li, Jun Peng, and Ze~Gao.
\newblock Inspire creativity with oriba: Transform artists' original characters
  into chatbots through large language model.
\newblock In \emph{Adjunct Proceedings of the 2023 ACM International Joint
  Conference on Pervasive and Ubiquitous Computing \& the 2023 ACM
  International Symposium on Wearable Computing}, pp.\  78--82,
  2023{\natexlab{b}}.

\bibitem[Swanson et~al.(2021)Swanson, Mathewson, Pietrzak, Chen, and
  Dinalescu]{swanson2021story}
Ben Swanson, Kory Mathewson, Ben Pietrzak, Sherol Chen, and Monica Dinalescu.
\newblock Story centaur: Large language model few shot learning as a creative
  writing tool.
\newblock In \emph{Proceedings of the 16th Conference of the European Chapter
  of the Association for Computational Linguistics: System Demonstrations},
  pp.\  244--256, 2021.

\bibitem[Tanaka et~al.(2022)Tanaka, Yamane, Mori, Mukuta, and
  Harada]{tanaka2022learning}
Kohtaro Tanaka, Hiroaki Yamane, Yusuke Mori, Yusuke Mukuta, and Tatsuya Harada.
\newblock Learning to evaluate humor in memes based on the incongruity theory.
\newblock In \emph{Proceedings of the Second Workshop on When Creative AI Meets
  Conversational AI}, pp.\  81--93, 2022.

\bibitem[Team(2024{\natexlab{a}})]{qvq-72b-preview}
Qwen Team.
\newblock Qvq: To see the world with wisdom, December 2024{\natexlab{a}}.
\newblock URL \url{https://qwenlm.github.io/blog/qvq-72b-preview/}.

\bibitem[Team(2024{\natexlab{b}})]{qwq-32b-preview}
Qwen Team.
\newblock Qwq: Reflect deeply on the boundaries of the unknown, November
  2024{\natexlab{b}}.
\newblock URL \url{https://qwenlm.github.io/blog/qwq-32b-preview/}.

\bibitem[Valitutti et~al.(2013)Valitutti, Toivonen, Doucet, and
  Toivanen]{valitutti2013let}
Alessandro Valitutti, Hannu Toivonen, Antoine Doucet, and Jukka~M Toivanen.
\newblock “let everything turn well in your wife”: generation of adult
  humor using lexical constraints.
\newblock In \emph{Proceedings of the 51st Annual Meeting of the Association
  for Computational Linguistics (Volume 2: Short Papers)}, pp.\  243--248,
  2013.

\bibitem[V{\'a}squez \& Aslan(2021)V{\'a}squez and Aslan]{vasquez2021cats}
Camilla V{\'a}squez and Erhan Aslan.
\newblock “cats be outside, how about meow”: multimodal humor and
  creativity in an internet meme.
\newblock \emph{Journal of Pragmatics}, 171:\penalty0 101--117, 2021.

\bibitem[Wang et~al.(2024)Wang, Bai, Tan, Wang, Fan, Bai, Chen, Liu, Wang, Ge,
  Fan, Dang, Du, Ren, Men, Liu, Zhou, Zhou, and Lin]{Qwen2VL}
Peng Wang, Shuai Bai, Sinan Tan, Shijie Wang, Zhihao Fan, Jinze Bai, Keqin
  Chen, Xuejing Liu, Jialin Wang, Wenbin Ge, Yang Fan, Kai Dang, Mengfei Du,
  Xuancheng Ren, Rui Men, Dayiheng Liu, Chang Zhou, Jingren Zhou, and Junyang
  Lin.
\newblock Qwen2-vl: Enhancing vision-language model's perception of the world
  at any resolution.
\newblock \emph{arXiv preprint arXiv:2409.12191}, 2024.

\bibitem[Wang et~al.(2023)Wang, Lv, Yu, Hong, Qi, Wang, Ji, Yang, Zhao, Song,
  et~al.]{wang2023cogvlm}
Weihan Wang, Qingsong Lv, Wenmeng Yu, Wenyi Hong, Ji~Qi, Yan Wang, Junhui Ji,
  Zhuoyi Yang, Lei Zhao, Xixuan Song, et~al.
\newblock Cogvlm: Visual expert for pretrained language models.
\newblock \emph{arXiv preprint arXiv:2311.03079}, 2023.

\bibitem[Wang et~al.(2022)Wang, Kordi, Mishra, Liu, Smith, Khashabi, and
  Hajishirzi]{wang2022self}
Yizhong Wang, Yeganeh Kordi, Swaroop Mishra, Alisa Liu, Noah~A Smith, Daniel
  Khashabi, and Hannaneh Hajishirzi.
\newblock Self-instruct: Aligning language models with self-generated
  instructions.
\newblock \emph{arXiv preprint arXiv:2212.10560}, 2022.

\bibitem[Wei et~al.(2022)Wei, Wang, Schuurmans, Bosma, Xia, Chi, Le, Zhou,
  et~al.]{wei2022chain}
Jason Wei, Xuezhi Wang, Dale Schuurmans, Maarten Bosma, Fei Xia, Ed~Chi, Quoc~V
  Le, Denny Zhou, et~al.
\newblock Chain-of-thought prompting elicits reasoning in large language
  models.
\newblock \emph{Advances in neural information processing systems},
  35:\penalty0 24824--24837, 2022.

\bibitem[Wu et~al.(2021)Wu, Lin, Yang, and Xu]{wu2021mumor}
Jiaming Wu, Hongfei Lin, Liang Yang, and Bo~Xu.
\newblock Mumor: A multimodal dataset for humor detection in conversations.
\newblock In \emph{Natural Language Processing and Chinese Computing: 10th CCF
  International Conference, NLPCC 2021, Qingdao, China, October 13--17, 2021,
  Proceedings, Part I 10}, pp.\  619--627. Springer, 2021.

\bibitem[Wu et~al.(2022)Wu, Jiang, Donsbach, Gray, Molina, Terry, and
  Cai]{wu2022promptchainer}
Tongshuang Wu, Ellen Jiang, Aaron Donsbach, Jeff Gray, Alejandra Molina,
  Michael Terry, and Carrie~J Cai.
\newblock Promptchainer: Chaining large language model prompts through visual
  programming.
\newblock In \emph{CHI Conference on Human Factors in Computing Systems
  Extended Abstracts}, pp.\  1--10, 2022.

\bibitem[Xie et~al.(2023)Xie, Zhang, Zhou, Li, Zhang, Hessel, Yang, and
  Liu]{xie2023funqa}
Binzhu Xie, Sicheng Zhang, Zitang Zhou, Bo~Li, Yuanhan Zhang, Jack Hessel,
  Jingkang Yang, and Ziwei Liu.
\newblock Funqa: Towards surprising video comprehension.
\newblock \emph{arXiv preprint arXiv:2306.14899}, 2023.

\bibitem[Xu et~al.(2023)Xu, Sun, Zheng, Geng, Zhao, Feng, Tao, and
  Jiang]{xu2023wizardlm}
Can Xu, Qingfeng Sun, Kai Zheng, Xiubo Geng, Pu~Zhao, Jiazhan Feng, Chongyang
  Tao, and Daxin Jiang.
\newblock Wizardlm: Empowering large language models to follow complex
  instructions.
\newblock \emph{arXiv preprint arXiv:2304.12244}, 2023.

\bibitem[Xu et~al.(2022)Xu, Liu, Liu, Li, Feng, Peng, Shi, Sun, and
  Wang]{xu2022hybrid}
Haojie Xu, Weifeng Liu, Jiangwei Liu, Mingzheng Li, Yu~Feng, Yasi Peng, Yunwei
  Shi, Xiao Sun, and Meng Wang.
\newblock Hybrid multimodal fusion for humor detection.
\newblock In \emph{Proceedings of the 3rd International on Multimodal Sentiment
  Analysis Workshop and Challenge}, pp.\  15--21, 2022.

\bibitem[Xu(2024)]{xu2024exploring}
Rongwu Xu.
\newblock Exploring chinese humor generation: A study on two-part allegorical
  sayings.
\newblock \emph{arXiv preprint arXiv:2403.10781}, 2024.

\bibitem[Xu et~al.(2024)Xu, Yuan, Chen, and Yang]{xu2024good}
Zhijun Xu, Siyu Yuan, Lingjie Chen, and Deqing Yang.
\newblock " a good pun is its own reword": Can large language models understand
  puns?
\newblock \emph{arXiv preprint arXiv:2404.13599}, 2024.

\bibitem[Yang et~al.(2024)Yang, Yang, Hui, Zheng, Yu, Zhou, Li, Li, Liu, Huang,
  Dong, Wei, Lin, Tang, Wang, Yang, Tu, Zhang, Ma, Xu, Zhou, Bai, He, Lin,
  Dang, Lu, Chen, Yang, Li, Xue, Ni, Zhang, Wang, Peng, Men, Gao, Lin, Wang,
  Bai, Tan, Zhu, Li, Liu, Ge, Deng, Zhou, Ren, Zhang, Wei, Ren, Fan, Yao,
  Zhang, Wan, Chu, Liu, Cui, Zhang, and Fan]{qwen2}
An~Yang, Baosong Yang, Binyuan Hui, Bo~Zheng, Bowen Yu, Chang Zhou, Chengpeng
  Li, Chengyuan Li, Dayiheng Liu, Fei Huang, Guanting Dong, Haoran Wei, Huan
  Lin, Jialong Tang, Jialin Wang, Jian Yang, Jianhong Tu, Jianwei Zhang,
  Jianxin Ma, Jin Xu, Jingren Zhou, Jinze Bai, Jinzheng He, Junyang Lin, Kai
  Dang, Keming Lu, Keqin Chen, Kexin Yang, Mei Li, Mingfeng Xue, Na~Ni, Pei
  Zhang, Peng Wang, Ru~Peng, Rui Men, Ruize Gao, Runji Lin, Shijie Wang, Shuai
  Bai, Sinan Tan, Tianhang Zhu, Tianhao Li, Tianyu Liu, Wenbin Ge, Xiaodong
  Deng, Xiaohuan Zhou, Xingzhang Ren, Xinyu Zhang, Xipin Wei, Xuancheng Ren,
  Yang Fan, Yang Yao, Yichang Zhang, Yu~Wan, Yunfei Chu, Yuqiong Liu, Zeyu Cui,
  Zhenru Zhang, and Zhihao Fan.
\newblock Qwen2 technical report.
\newblock \emph{arXiv preprint arXiv:2407.10671}, 2024.

\bibitem[Yang et~al.(2023)Yang, Huang, Xia, and Huang]{yang2023knowledge}
Yuhao Yang, Chao Huang, Lianghao Xia, and Chunzhen Huang.
\newblock Knowledge graph self-supervised rationalization for recommendation.
\newblock In \emph{Proceedings of the 29th ACM SIGKDD conference on knowledge
  discovery and data mining}, pp.\  3046--3056, 2023.

\bibitem[Yao et~al.(2024)Yao, Yu, Zhao, Shafran, Griffiths, Cao, and
  Narasimhan]{yao2024tree}
Shunyu Yao, Dian Yu, Jeffrey Zhao, Izhak Shafran, Tom Griffiths, Yuan Cao, and
  Karthik Narasimhan.
\newblock Tree of thoughts: Deliberate problem solving with large language
  models.
\newblock \emph{Advances in Neural Information Processing Systems}, 36, 2024.

\bibitem[Ye et~al.(2023)Ye, Xu, Xu, Ye, Yan, Zhou, Wang, Hu, Shi, Shi,
  et~al.]{ye2023mplug}
Qinghao Ye, Haiyang Xu, Guohai Xu, Jiabo Ye, Ming Yan, Yiyang Zhou, Junyang
  Wang, Anwen Hu, Pengcheng Shi, Yaya Shi, et~al.
\newblock mplug-owl: Modularization empowers large language models with
  multimodality.
\newblock \emph{arXiv preprint arXiv:2304.14178}, 2023.

\bibitem[Zeng et~al.(2022)Zeng, Attarian, Ichter, Choromanski, Wong, Welker,
  Tombari, Purohit, Ryoo, Sindhwani, et~al.]{zeng2022socratic}
Andy Zeng, Maria Attarian, Brian Ichter, Krzysztof Choromanski, Adrian Wong,
  Stefan Welker, Federico Tombari, Aveek Purohit, Michael Ryoo, Vikas
  Sindhwani, et~al.
\newblock Socratic models: Composing zero-shot multimodal reasoning with
  language.
\newblock \emph{arXiv preprint arXiv:2204.00598}, 2022.

\bibitem[Zeng et~al.(2024)Zeng, Xu, Zhao, Lou, and Chen]{zeng2024automatic}
Weihao Zeng, Can Xu, Yingxiu Zhao, Jian-Guang Lou, and Weizhu Chen.
\newblock Automatic instruction evolving for large language models.
\newblock \emph{arXiv preprint arXiv:2406.00770}, 2024.

\bibitem[Zhang et~al.(2020)Zhang, Liu, Lv, and Luo]{zhang2020let}
Hang Zhang, Dayiheng Liu, Jiancheng Lv, and Cheng Luo.
\newblock Let's be humorous: Knowledge enhanced humor generation.
\newblock \emph{arXiv preprint arXiv:2004.13317}, 2020.

\bibitem[Zhang et~al.(2022)Zhang, Zhang, Li, and Smola]{zhang2022automatic}
Zhuosheng Zhang, Aston Zhang, Mu~Li, and Alex Smola.
\newblock Automatic chain of thought prompting in large language models.
\newblock \emph{arXiv preprint arXiv:2210.03493}, 2022.

\bibitem[Zhong et~al.(2024)Zhong, Huang, Gao, Wen, Lin, Zitnik, and
  Zhou]{zhong2024let}
Shanshan Zhong, Zhongzhan Huang, Shanghua Gao, Wushao Wen, Liang Lin, Marinka
  Zitnik, and Pan Zhou.
\newblock Let's think outside the box: Exploring leap-of-thought in large
  language models with creative humor generation.
\newblock In \emph{Proceedings of the IEEE/CVF Conference on Computer Vision
  and Pattern Recognition}, pp.\  13246--13257, 2024.

\bibitem[Zhou et~al.(2023)Zhou, Lu, Mishra, Brahma, Basu, Luan, Zhou, and
  Hou]{zhou2023instruction}
Jeffrey Zhou, Tianjian Lu, Swaroop Mishra, Siddhartha Brahma, Sujoy Basu,
  Yi~Luan, Denny Zhou, and Le~Hou.
\newblock Instruction-following evaluation for large language models.
\newblock \emph{arXiv preprint arXiv:2311.07911}, 2023.

\end{thebibliography}
